\documentclass[letterpaper]{article} % DO NOT CHANGE THIS
\usepackage{aaai2026}  % DO NOT CHANGE THIS
\usepackage{times}  % DO NOT CHANGE THIS
\usepackage{helvet}  % DO NOT CHANGE THIS
\usepackage{courier}  % DO NOT CHANGE THIS
\usepackage[hyphens]{url}  % DO NOT CHANGE THIS
\usepackage{graphicx} % DO NOT CHANGE THIS
\usepackage{subcaption}
\usepackage{natbib}  % DO NOT CHANGE THIS AND DO NOT ADD ANY OPTIONS TO IT
\usepackage{caption} % DO NOT CHANGE THIS AND DO NOT ADD ANY OPTIONS TO IT
\usepackage{algorithm}
\usepackage{algorithmic}
\usepackage{multirow} 
\usepackage{booktabs}

\usepackage{newfloat}
\usepackage[utf8]{inputenc}

\usepackage{amsmath,amsfonts}
\usepackage{dsfont}
\newtheorem{definition}{Definition}
\usepackage{bm}

\usepackage{listings}
\DeclareCaptionStyle{ruled}{labelfont=normalfont,labelsep=colon,strut=off} % DO NOT CHANGE THIS
\floatstyle{ruled}
\newfloat{listing}{tb}{lst}{}
\floatname{listing}{Listing}
\title{MTP: Exploring Multimodal Urban Traffic Profiling with Modality \\  Augmentation  and Spectrum Fusion}
\author{
   Haolong Xiang\textsuperscript{\rm 1, \rm 2},
   Peisi Wang\textsuperscript{\rm 1},
    Xiaolong Xu\textsuperscript{\rm 1},
    Kun Yi\textsuperscript{\rm 3},
    Xuyun Zhang\textsuperscript{\rm 4},
    Quanzheng Sheng\textsuperscript{\rm 4},\\
    Amin Beheshti\textsuperscript{\rm 4},
    Wei Fan\textsuperscript{\rm 5}\thanks{Corresponding author.}
}
\affiliations{
    \textsuperscript{\rm 1} School of Software, Nanjing University of Information Science and Technology, Nanjing, P.R. China\\
    \textsuperscript{\rm 2}State Key Lab. for Novel Software Technology, Nanjing University, P.R. China\\  
    \textsuperscript{\rm 3}State Information Center, P.R. China\\
    \textsuperscript{\rm 4}School of  Computing, Macquarie University, NSW, Australia\\
     \textsuperscript{\rm 5}School of Computer Science, University of Auckland, Auckland, New Zealand\\
    
    \{hlxiang, 202512492293, xlxu\}@nuist.edu.cn, yikun@bit.edu.cn, \{xuyun.zhang, michael.sheng, amin.beheshti\}@mq.edu.au, wei.fan@auckland.ac.nz
}

\usepackage{bibentry}
\begin{document}

\maketitle

\begin{abstract}
With rapid urbanization in the modern era, traffic signals from various sensors have been playing a significant role in monitoring the states of cities, which provides a strong foundation in ensuring safe travel, reducing traffic congestion and optimizing urban mobility. Most existing methods for traffic signal modeling often rely on the original data modality, i.e., numerical direct readings from the sensors in cities. However, this unimodal approach overlooks the semantic information existing in multimodal heterogeneous urban data in different perspectives, which hinders a comprehensive understanding of traffic signals and limits the accurate prediction of complex traffic dynamics. To address this problem, we propose a novel \textit{M}ultimodal framework, \textit{MTP}, for urban \textit{T}raffic \textit{P}rofiling, which learns multimodal features through numeric, visual, and textual perspectives. 
%The three branches drive for a multimodal perspective of traffic signal learning for , while the frequency learning strategies delicately refine the information for extraction. 
The three branches drive for a multimodal perspective of urban traffic signal learning in the frequency domain, while the frequency learning strategies delicately refine the information for extraction. Specifically, we first conduct the visual augmentation for the traffic signals, which transforms the original modality into frequency images and periodicity images for visual learning. Also, we augment descriptive texts for the traffic signals based on the specific topic, background information and item description for textual learning. To complement the numeric information, we utilize frequency multilayer perceptrons for learning on the original modality. We design a hierarchical contrastive learning on the three branches to fuse the spectrum of three modalities. Finally, extensive experiments on six real-world datasets demonstrate superior performance compared with the state-of-the-art approaches. 
\end{abstract}

\begin{links}
    \link{Code}{https://github.com/jorcy3/MTP}
    % \link{Extended version}{https://aaai.org/example/extended-version}
\end{links}

\section{Introduction}
With rapid urbanization, traffic volumes continue to rise, placing unprecedented pressure on transportation systems \cite{zhou2025trafficformer,wu2025big}. Persistent congestion during peak hours, delayed responses to traffic incidents, and imbalanced road network resource allocation not only significantly harm the travel efficiency and experience of citizens but also directly restrict the operational efficiency of urban economies and environmental sustainability \cite{liu2025multi}. As a core carrier reflecting how transportation systems operate, traffic time series data contains critical information such as traffic flow variations, the operation status of road segments, and early signs of abnormal events \cite{fang2025efficient}. Thorough profiling of these data enables real-time perception and scientific assessment of traffic conditions. It can quickly identify congested road segments and their congestion levels to provide precise guidance for traffic management authorities \cite{wang2024moderntcn}. Additionally, it offers data support for urban road network planning, the optimization of public transportation routes, and the adjustment of traffic signal timing, thereby fundamentally enhancing the operational efficiency of transportation systems \cite{xiang2025Empower}. Currently, urban traffic profiling is a fundamental component for achieving intelligent traffic management and addressing urban traffic challenges. 

Traditional traffic data processing methods mostly rely on static feature extraction, such as sliding window statistics and support vector machines, but they assume that the data distribution is stable and cannot adapt to the dynamic characteristics of actual traffic systems \cite{zerveas2021tst}. In practice, traffic data has strong temporal dependence, i.e., the traffic flow characteristics of main roads during morning peak hours are significantly different from those during off-peak hours, and the traffic status under extreme weather may even deviate from the conventional distribution \cite{cheng2021shapenet}. This temporal dynamics in traffic signals leads to a sharp decline in the accuracy of static methods in cross-time profiling \cite{zhang2023lightts}. For example, a model trained based on morning peak data will misjudge normal traffic flow during off-peak hours as abnormal. Therefore, designing a dedicated profiling method for the comprehensive temporal features has become a fundamental research issue for the accurate classification of traffic states.

Urban traffic profiling can be divided into two core tasks: one is state profiling, such as smooth, slow, and congested, and the other is event profiling, such as traffic accidents and road construction \cite{nie2023patchtst}. Neural networks have become the mainstream method due to their time series modeling capabilities, e.g., SVP-T \cite{zhou2023svpt} learns representations from both the shape-level and velocity-level of the time series for more robust feature capture. However, traffic data has expanded from single structured time series data to a multimodal form, such as surveillance images, text information, and social media feedback have been incorporated into the analysis \cite{wen2025interpgn}. Although these data can supplement semantic information, traditional neural networks are designed for a single modality and cannot achieve deep correlation between spatial, visual and textual features, resulting in limited utilization of multimodal information.

To meet the needs of multimodality, large language models (LLMs) based methods have gradually developed for better urban traffic profiling. In terms of time series modeling, TRACK \cite{han2025bridging} leverages transformer-based models to learn dynamic road network and trajectory representations for better capturing spatial-temporal dynamics. Besides, CAFO \cite{li2024cafo} effectively combines the local feature extraction capabilities of convolutional layers with the ability of attention mechanisms to capture long-range dependencies. In terms of multimodal fusion, urban-level CLIP \cite{yan2024urbanclip} realizes the associated classification of urban images and texts through visual-text pre-training. Although the above methods have made certain progress, they still face core challenges in actual traffic scenarios: existing LLMs are mostly optimized for a single modality. \textit{Large large models} are good at processing image data but are hard to parse the dynamic changes of time series features~\cite{gruver2023large}. \textit{Text-Augmented Models} can understand traffic event descriptions but lack the ability to model the time dimension~\cite{wang2024data}. \textit{Time series large models} cannot effectively integrate semantic information in images and texts~\cite{wang2024deep}. Despite advancements in textual and visual large models, it has been less investigated in traffic classification by integrating multiple modalities. 

% The above features drive for a multimodal perspective of urban traffic signal learning in the frequency domain, while the frequency learning strategies delicately refine the information for extraction. Furthermore, we design a hierarchical contrastive learning and feature fusion scheme on the three branches for better representation learning in traffic profiling tasks.
To address the above issues, we propose a new Multimodal framework, \textit{MTP}, for urban Traffic Profiling. Specifically, MTP first augments visual and text traffic profiles using the original traffic signals and then incorporates multiple features, including temporal, visual, and textual information for learning. The main contributions are: %of our work are three-fold:
\begin{itemize}
	\item We propose a novel multimodal framework for urban traffic profiling, which firstly augments multimodal features on traffic signals and learns through numeric, visual, and textual perspectives in the frequency domain. 
	\item  We design a hierarchical contrastive learning on the augmented image, text, and numerical value to optimize the multimodal learning and fuse the three representations.
	\item Extensive experiments are conducted on six real-world datasets, which validates the effectiveness of the proposed framework compared with the state-of-the-art baselines. We also design several ablation studies to show the influence of three different modalities and conduct qualitative analysis with visualization for our framework. 
\end{itemize}

\section{Related Work}
% This section reviews three research areas most relevant to our work: traditional time series traffic profiling, time series analysis based on Large Language Models (LLMs), and traffic analysis using Vision-Language Models (VLMs). By examining the achievements and limitations of existing work, we aim to identify the current research gap and position the unique contribution of our work.

\noindent \textbf{Traditional Traffic Time Series Profiling.}
Existing methods for road traffic condition analysis mostly rely on single-modality data. In the field of time series analysis, deep learning techniques such as Convolutional Neural Networks (CNNs) \cite{he2016deep, alam2023feature}, Recurrent Neural Networks (RNNs) \cite{jin2017multimodal, zheng2020learning}, Graph Neural Networks (GNNs) \cite{zhang2023tfe, deng2024annet}, and Transformer-based methods \cite{lin2022et, zerveas2021tst} have been widely used to analyze various traffic conditions. These methods excel at processing structured time series data, driver profiling \cite{cura2020driver}, and assessing driving risks \cite{abdelrahman2020robust}. However, their core limitation lies in their unimodal nature. Merely analyzing time series data or isolatedly analyzing image and text information is insufficient to capture dynamic real-world traffic conditions.

\noindent \textbf{Traffic Profiling with LLMs.}
LLMs have powerful capabilities in processing multimodal data, especially in text understanding and generalization \cite{khattar2019mvae, feng2024cp}, which offer new avenues to address the problem of single-modality information loss. In recent years, researchers have begun to apply LLMs to the field of intelligent transportation. For example, the multimodal framework proposed by \citet{qian2021hierarchical} combines BERT and ResNet to jointly capture contextual information; \citet{chen2024ilm} utilize an LLM-driven framework to optimize vehicle dispatching and navigation; and \citet{yan2024urbanclip} use LLMs to enhance textual information and fuse it with images via contrastive learning to generate multimodal representations for urban region profiling. Although these methods demonstrate the potential of LLMs, their applications are often task-specific, which partially hinders their exploration in more general road traffic profiling research.

\noindent \textbf{Traffic Profiling with VLMs.}
VLMs have made significant breakthroughs in jointly processing and understanding visual and textual information. Many recent works, such as BLIVA \cite{hu2024bliva}, EMMA \cite{yang2024embodied}, and OmniActions \cite{li2024omniactions}, have demonstrated the powerful capabilities of VLMs in handling complex visual question answering and multimodal interaction tasks. However, these methods have not fully combined multimodal data to generate powerful representations for road traffic profiling. These methods indicate that VLMs can serve as a powerful ``bridge'' to transform visual information into high-quality textual information, laying the foundation for subsequent multimodal fusion.

Despite significant progress in the fields mentioned above, a key research gap remains: The joint modeling and fusion of numerical, textual and visual modalities have not been explored in urban traffic profiling, which largely hinders the accurate prediction or classification of traffic conditions.

% Specifically, current research lacks a unified framework that can systematically integrate traditional time series analysis, LLM-driven text generation, and VLM-driven visual information transformation. To address this issue, we propose \texttt{MTP}, an innovative multimodal learning and fusion framework for road traffic profiling. To the best of our knowledge, this is the first attempt of its kind in this field. Our work aims to integrate information from time series, text, and image dimensions in an end-to-end manner, providing a more comprehensive and powerful solution for understanding and analyzing complex traffic conditions.

\section{Problem Definition}
\begin{definition}[Urban Area]
Given an urban area $\mathbb{U}$, we can divide it into $M$ traffic jurisdictions. For different time intervals $T$, a corresponding traffic status profiling is conducted.
\end{definition}

\begin{definition}[Numerical representation]
By using devices such as sensors or cameras, we can collect data information within an urban area $\mathbb{U}$. Then, the data of urban traffic will be stored in numerical representation, denoted as $\bm{v_\mathbb{U}}$, which contains information such as the traffic background, vehicle position, environment, and item description.
\end{definition}

\begin{definition}[Image Augmented representation]
The image data is generated from the original traffic time series data, denoted as $\bm{g_\mathbb{U}}$, which enables the model to capture spatial features from the original temporal data.
\end{definition}

\begin{definition}[Text Augmented representation]
The text data is generated from the original urban traffic data, denoted as $\bm{t_\mathbb{U}}$, which enables the model to capture semantic information and from the original temporal data.
\end{definition}

This paper mainly deals with urban traffic analysis that focuses on traffic road conditions and vehicle flow, which we define as a classification task. The state information of the data includes three modalities: original numerical values $v$, images $g$, and texts $t$. Given a traffic time series dataset $\bm{X}={(\bm{x_1},y_1), (\bm{x_2}, y_2),...,(\bm{x_n}, y_n)}$, where each data instance $(\bm{x_i},y_i)$ contains the feature representation $\bm{x_i}$ and the types of transportation $y_i$. Specifically, the feature representation can be optimized through our multi-modal feature fusion, which can be calculated by:
\begin{equation}
\bm{x^{\prime}_\mathbb{U}} = \mathbb{H} (\bm{v_\mathbb{U}},\bm{g_\mathbb{U}},\bm{t_\mathbb{U}}|\bm{x}).
\end{equation}

\noindent Finally, we can use the augmented representation to predict the traffic condition through $\bm{x^{\prime}_\mathbb{U}} \rightarrow y_i$.

\section{Methodology}
% We propose a new multimodal framework (MTP) for urban traffic profiling, which fine-tunes the LLMs with multi-features, including temporal, visual, and textual information. As shown in Figure ~\ref{fig:framework}, MTP consists of three modal encoder branches and a feature fusion scheme: a) time series modality encoder, b) vision modality encoder and c) text modality encoder. The first branch processes the raw traffic time series data, with the core being the extraction of time series features through frequency-domain learning. The visual branch converts the time series into a visual modality for learning, and then performs feature noise reduction through spectral conversion technology. The textual branch generates descriptive texts based on time series and extract semantic features, and then obtains more accurate representations through feature fusion processing of vision and text. Finally, the above three branches are integrated through the fusion module by cross spectrum fusion and hierarchical contrastive learning. The fused features of our framework can simultaneously retain the temporal patterns of the numerical modality, the intuitive patterns of the visual modality, and the semantic information of the text modality.

\begin{figure*}[t!]
    \centering
    \includegraphics[width=0.97\linewidth]{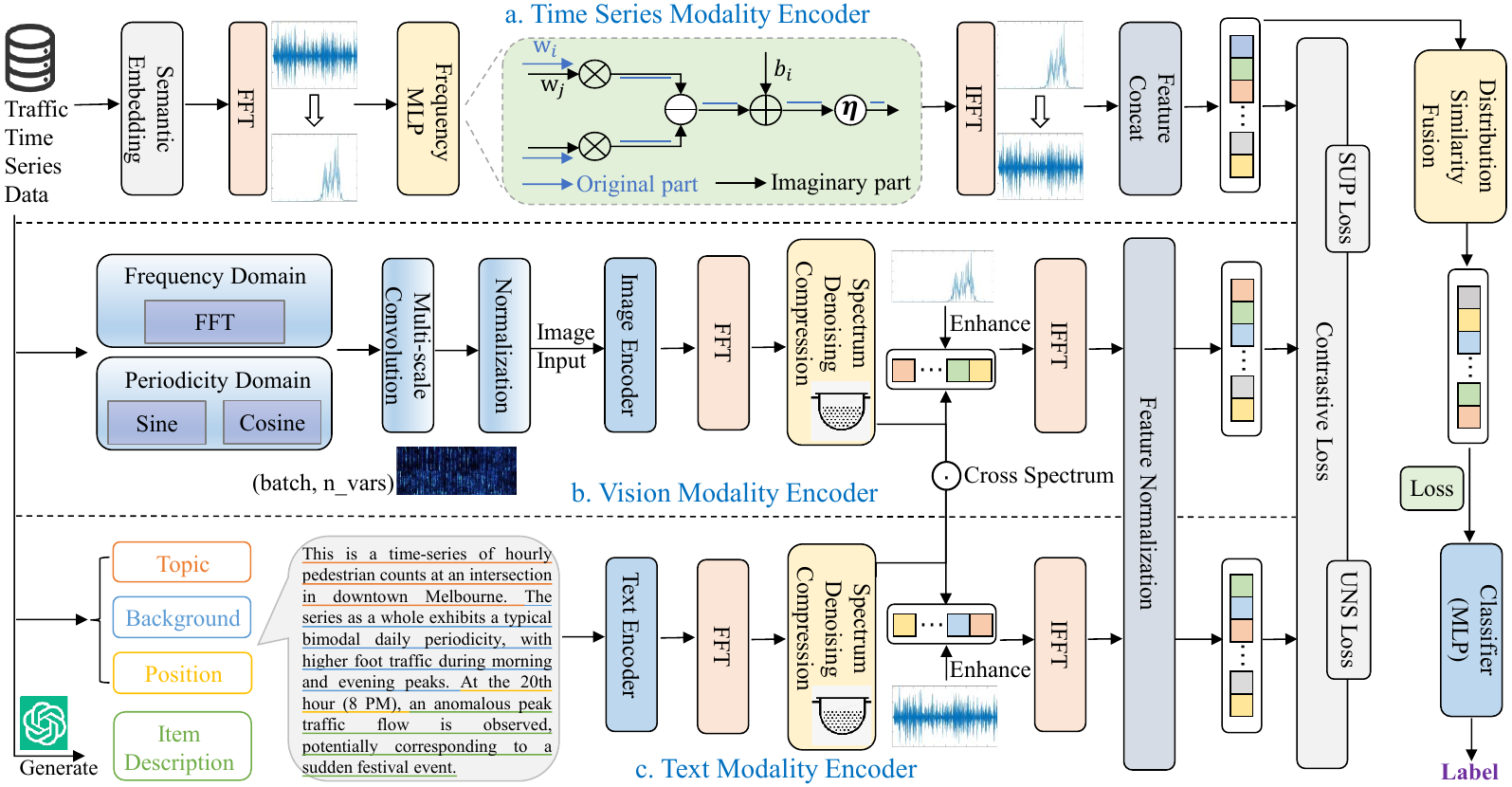}
    \caption{The overview of our framework. MTP learns multimodal features in the frequency domain from three perspectives: numerical, visual, and textual. These modalities are fused to provide more comprehensive features for urban traffic profiling. 
}
    \label{fig:framework}
\end{figure*}

We propose a novel multimodal framework (MTP) for urban traffic profiling, which learns multimodal features through numeric, visual, and textual perspectives in the frequency domain. As shown in Figure ~\ref{fig:framework}, MTP consists of three modal encoder branches and a feature fusion scheme: a) time series modality encoder, b) vision modality encoder and c) text modality encoder. The fused features of our framework can simultaneously retain the temporal patterns of the numerical modality, the intuitive patterns of the visual modality, and the semantic information of the text modality. The following sections will elaborate on the design of each module of our framework.

\subsection{Time Series Modality Encoder}
This module mainly processes the original traffic time series data modality with spectrum conversion technologies, as shown in ``a'' part of Figure \ref{fig:framework}. Time series modality encoder mainly involves the semantic embedding, Fast Fourier Transform (FFT), frequency-domain multi-layer perceptions (MLPs), and inverse Fast Fourier Transform (IFFT). Inspired by word embeddings ~\cite{mikolov2013efficient}, we mapped the input data $\bm{X} \in \mathbb{R}^{n \times l}$ into a hidden representation $\bm{D} \in \mathbb{R}^{n \times l \times m}$ to introduce richer semantic information, which is realized by a learnable weight vector $\bm{\psi} \in \mathbb{R}^{1 \times m}$. The process can be denoted as $\bm{D} = \bm{X} \times \bm{\psi}$.

The second step is to convert the spatial domain to the frequency domain, so that our model can extract multi-scale features and periodic features of the traffic time series data. Given the converted input $\bm{D}$, the Fourier transform of the original time series embedding is defined as:
\begin{equation}
\bm{\mathcal{D}}^v[k] = \sum_{i=0}^{n-1}\bm{D}^v[i]e^{-j\frac{2\pi ki}{n}},
\end{equation}
\noindent where $i$ represents the integral variable, $j$ represents the imaginary unit, and $e^{-j\frac{2\pi ki}{n}} = \cos(\frac{2\pi ki}{n}) - j \sin (\frac{2\pi ki}{n})$. Through the above process, we can obtain the numerical spectrum at the frequency $2\pi ki/n$.

The obtained frequency components are input into the frequency-domain MLPs, and operations are performed through the set complex weight matrix $\bm{W}$ and bias $\bm{B}$ to obtain the frequency-domain output result:
\begin{equation}
\bm{\mathcal{H}}_i = FMLP(\bm{\mathcal{D}}^v,\bm{W},\bm{B}).
\end{equation}
\noindent The concrete process of frequency-domain MLPs is shown in the green box of the framework figure. The core function of frequency-domain MLPs is to perform nonlinear mapping and feature extraction on the frequency domain features after FFT conversion, enhancing the expression ability of the frequency domain features to meet the requirements for capturing periodic and abnormal patterns in traffic time series analysis. Specifically, frequency-domain MLPs can be calculated by:
\begin{equation}
\bm{\mathcal{Z}} = ReLU(\mathcal{H} \bm{W}+\bm{B}).
\label{Equ:mlp}
\end{equation}
\noindent If the MLPs consists of $l$ layers, then the input of each layer is the output ($\bm{\mathcal{Z}^l}$) of the frequency-domain MLPs of the previous layer. The complex weight matrix $\bm{W}$ fulfill the condition: $\bm{W}=\bm{W}_i+ \eta\bm{W}_j$, and bias $\bm{B}$ fulfill the condition: $\bm{B}=\bm{B}_i+ \eta\bm{B}_j$. According to the rule of multiplication of complex numbers, we can derive the following condition from Equation (\ref{Equ:mlp}):
\begin{equation}
\begin{aligned}
\bm{\mathcal{Z}}^l &= ReLU(O(\mathcal{Z}^{l-1}) \bm{W}^l_i-I(\mathcal{Z}^{l-1}) \bm{W}^l_j+\bm{B}^l_i) \\
& + \eta ReLU(O(\mathcal{Z}^{l-1}) \bm{W}^l_j-I(\mathcal{Z}^{l-1}) \bm{W}^l_i+\bm{B}^l_j).
\end{aligned}
\end{equation}
\noindent where $O$ represents the original parts of frequency components, and $I$ represents the imaginary parts of frequency components.

We use Inverse Fast Fourier Transform (IFFT) to inverse the optimized frequency-domain features to the spatial domain, which provides the features with frequency-domain information. This process provide feature support in the spatial domain for the subsequent feature concatenation (Concat), similarity calculation, and traffic series fusion. The calculation of IFFT is formularized as:
\begin{equation}
\bm{D}^v[i] = \sum_{i=0}^{n-1} \bm{\mathcal{D}}^v[k]e^{j\frac{2\pi ki}{n}},
\end{equation}

\subsection{Vision Modality Encoder}
The ``b'' part in Figure~\ref{fig:framework} is a feature extraction module for the visual modality. The process of the vision modality encoder is to first convert the traffic time series data into an image, and then perform frequency-domain processing on the image to extract visual features. In our approach, the essence of image generation is to convert the spatial-domain traffic time series data into visual images, achieving the transformation from numerical modality to visual modality. Specifically, we applied the FFT to extract frequency information from the input data as the frequency domain encoder. The extracted frequency representations are concatenated with the original input. Besides, we also design the periodicity domain encoder to extract the temporal dependencies. For each time stamp $t$, we use the following equation to get the new encoder:
\begin{equation}
\bm{P}_t = [\sin(2 \pi t/ \phi),\cos(2 \pi t/ \phi)],
\end{equation}
\noindent where $\phi$ represents the periodicity hyperparameter. These encodings are also concatenated with the original input, which constitutes a group of new representations $\bm{X}^g$.

Next, we employ multi-scale convolution to extract hierarchical temporal patterns. Specifically, we first use a 1D convolutional layer to capture local dependencies. Among the subsequent two 2D convolutional layers, one halves the channel dimension, and another maps features to multiple output channels, thereby capturing global temporal structures. The output features are resized to the desired image dimensions via bilinear interpolation and then subjected to normalization.

After generating the image dimensions, we use the image encoder to obtain the numerical representations. These features are also converted into the frequency domain by FFT, which can be formulated as:
\begin{equation}
\bm{\mathcal{X}}^g[k] = \sum_{i=0}^{n-1}\bm{X}^g[i]e^{-j\frac{2\pi ki}{n}}.
\end{equation}

In order to reduce the noise contained in the augmented images and focus on the core information, we introduced the finite impulse response (FIR) filter to process the features in the frequency domain. The FIR filter is constructed based on the ``window'' technique, specifically using the Hamming window. The characteristics of the Hamming window allow the filter to naturally aggregate the main information in traffic data while smoothing spectrum fluctuations. Given the filter length $s$, we can generate window function parameters through the Hamming window function by:
\begin{equation}
\omega_i=0.54-0.46\cos(z\pi i/s-1).
\end{equation}
\noindent Then, we can get the actual impulse response $r^i = \omega[i] \cdot r^{\prime}$ by multiplying the window function with the filter's ideal impulse response $r^{\prime}$. These impulse responses form a filter bank $\bm{R} = [r^1, r^2, ..., r^s]$ with $s$ filters. The filter bank can divide the input spectrum into multiple sub-bands. Through the impulse response $r^i$ of each filter, key features within the corresponding frequency range are filtered out to avoid interference from irrelevant frequencies. Through the spectrum compression, we can calculate the spectrum by:
\begin{equation}
\bm{\mathcal{X}}^g_{spe} = \sum_{i=1}^s \frac{1}{c}  |\bm{\mathcal{X}}^g|^2 \odot r^i,
\end{equation}
\noindent where $c$ represents the length of image modality, and $\odot$ represents the element-wise multiplication. Essentially, the filter is used to weighted the spectrum through this operation, which retains important frequency components and weakens redundant information, ultimately achieving efficient spectrum compression.

To address the limitation of fixed spectrum compression in being unable to remove high-frequency noise in traffic image processing, average pooling is introduced. It reduces high-frequency noise and random fluctuations by smoothing the spectrum, preserves the overall trend to make the spectrum more regular, and thereby improves the efficiency of compression algorithms and the effect of traffic image feature recognition. The average pooling process can be calculated by:
\begin{equation}
\bm{\mathcal{X}}_{pool}^g = Average(\bm{\mathcal{X}}^g_{spe}\odot \bm{\delta}^g),
\end{equation}
\noindent where $\bm{\delta}^g$ represents a matrix that holds the corresponding dimension with $\bm{X}^g_{spe}$. In terms of cross-modal fusion, the enhanced spectrum of images is generated with the help of text modal information, and the formula is:
\begin{equation}
\bm{\mathcal{X}}_{out}^g = \bm{\mathcal{X}}^g_{spe} \odot \bm{\mathcal{X}}_{pool}^t,
\end{equation}
\noindent where $\bm{X}_{pool}^t$ represents the output of text modality with pooling enhancement.

After pooling enhancement and spectrum cross in the frequency domain, we apply the IFFT to invert the features to the spatial representations. The IFFT of image features can be calculated by:
\begin{equation}
\bm{X}^g[i] = \sum_{i=0}^{n-1} \bm{\mathcal{X}}_{out}^g[k]e^{j\frac{2\pi ki}{n}}.
\end{equation}

\subsection{Text Modality Encoder}

In this module, text can be pre-defined in the original traffic time series data or generated from the input data. If a generated textual description is required, we design text generation standards shown in  ``c'' part of Figure \ref{fig:framework}. First, we can use LLMs (e.g., ChatGPT) to generate some item descriptions, which can enhance the semantic information for textual feature extraction. Then, more contextual information, such as topic, background, and vehicle position, can be extracted directly from the input data, thereby facilitating complete textual information for traffic profiling. If the input data already contains complete textual information, it can be directly fed into the text encoder to generate the vector features for subsequent processing.

Similar to the previous vision modality encoder, we use spectrum transformation technology to convert the vector generated by the text encoder into the frequency domain, followed by denoising and cross-modal spectrum processing with the image modality. Given the vector $\bm{X}^t$ generated by the text encoder, the representation in the frequency domain can be formulated as:
\begin{equation}
\bm{\mathcal{X}}^t[k] = \sum_{i=0}^{n-1}\bm{X}^t[i]e^{-j\frac{2\pi ki}{n}}.
\end{equation}

Through FIR filter, average pooling, and cross-modal enhancement processing, we can calculate new spectral representations by:
\begin{equation}
\bm{\mathcal{X}}_{out}^t = \bm{\mathcal{X}}^t_{spe} \odot Average(\bm{\mathcal{X}}^g_{spe}\odot \bm{\delta}^g).
\end{equation}

Finally, IFFT is applied to invert the frequency-domain features into the spatial-domain features $\bm{\mathcal{X}}_{out}^t$ for further cross-modal fusion.

\subsection{Cross-modal Fusion}
After each modality undergoes spectral transformation and frequency domain processing, the feature fusion is achieved through two schemes: contrastive learning and distribution similarity fusion.

\textbf{Contrastive Learning.} In our framework, the significance of contrastive loss lies in achieving semantic alignment of cross-modal features by reducing the distance between different modal features of the same traffic scene, while increasing the distance between irrelevant modal features, thereby enhancing the consistency of multi-modal features. For the labeled data, we can first conduct supervised learning to learn the supervised loss $\mathcal{L}(SUP)$. Given a data instance $x_i$, we can get the pairwise $(\bm{x_i^{\prime}},\bm{s_i})$ to calculate the supervised loss, where $\bm{x_i^{\prime}}$ corresponds to the encoding feature and $\bm{s_i}$ corresponds to the real feature. Given a dataset with $m$ categories, we can divide all instances into these $m$ types $\bm{\mathcal{Y}}=\{ \bm{\mathcal{M}}_1,\bm{\mathcal{M}}_2,...,\bm{\mathcal{M}}_m\}$. For each instance, we can define a supervised loss as $\mathcal{L}_i(\bm{x^{prime}},\bm{s_i})$ \cite{lin2022modeling}. Next, the whole supervised loss is calculated by:
\begin{equation}
\begin{aligned}
\mathcal{L}(SUP) &= \sum_{\bm{X}} \sum_{\bm{\mathcal{Y}}} (\sum_{\bm{x}^{\prime} \in \bm{\mathcal{M}}_i} \frac{1}{|\bm{\mathcal{M}}_i|} \sum_{\bm{s} \in \bm{\mathcal{M}}_i, \bm{x}^{\prime}\neq \bm{s}} [\mathcal{L}_i(\bm{x}^{\prime v},\bm{s}^v) \\
&+\mathcal{L}_i(\bm{x}^{\prime g},\bm{s}^g)+\mathcal{L}_i(\bm{x}^{\prime t},\bm{s}^t)]).
\end{aligned}
\end{equation}

Unsupervised learning mainly captures the differences between modalities by aligning the features of different modalities. We introduce the InfoNCE loss \cite{he2020momentum} to calculate the similarity, which is defined as follows:
\begin{equation}
\begin{aligned}
\mathcal{L}(UNS) &= \frac{1}{3|\bm{X}|}\sum_{i=1}^{|\bm{X}|} [\mathcal{L}_v (\bm{x}_i^v,\bm{x}_i^g,\bm{x}_i^t)+\mathcal{L}_g (\bm{x}_i^g,\bm{x}_i^t,\bm{x}_i^v) \\
& +\mathcal{L}_t (\bm{x}_i^t,\bm{x}_i^v,\bm{x}_i^g)].
\end{aligned}
\end{equation}

\textbf{Fusion Loss.} To ensure the semantic consistency of cross-modal features, we design a distribution similarity fusion scheme to assess the similarity between different modal features. Specifically, we apply Jensen-Shannon (JS) divergence between any two modalities to calculate the distribution similarity. Given a data instance $x$, its posterior probability in numerical modality can be defined as $\mathbb{I}(\bm{\alpha}^v|\bm{x}^v)$. After distribution similarity fusion, the JS divergence can be calculated by:
\begin{equation}
\begin{aligned}
\Delta &=(JS(\mathbb{I}(\bm{\alpha}^v|\bm{x}^v) || \mathbb{I}(\bm{\alpha}^g|\bm{x}^g)) + JS(\mathbb{I}(\bm{\alpha}^v|\bm{x}^v) || \mathbb{I}(\bm{\alpha}^t|\bm{x}^t)) \\
& + JS(\mathbb{I}(\bm{\alpha}^g|\bm{x}^g) || \mathbb{I}(\bm{\alpha}^t|\bm{x}^t)))/3,
\end{aligned}
\end{equation}

\begin{table*}[ht]
\centering
\scriptsize % 使用允许的最小字体以适应极宽的表格
\setlength{\tabcolsep}{3pt} % 极限压缩列间距
\caption{Overall performance comparison on all datasets. Our proposed model (MTP) is compared with state-of-the-art baselines on metrics F1-score (F1) and accuracy (Acc). The best result is in \textbf{bold}, and the second-best is \underline{underlined}.}
\label{tab:sota_final}
\begin{tabular}{l cc cc cc cc cc cc cc cc cc}
\toprule
\multirow{2}{*}{Dataset} & \multicolumn{2}{c}{ShapeNet} & \multicolumn{2}{c}{TST} & \multicolumn{2}{c}{PatchTST} & \multicolumn{2}{c}{SVP-T} & \multicolumn{2}{c}{LightTS} & \multicolumn{2}{c}{ModernTCN} & \multicolumn{2}{c}{CAFO} & \multicolumn{2}{c}{InterpGN} & \multicolumn{2}{c}{\textbf{MTP}} \\
\cmidrule(lr){2-3} \cmidrule(lr){4-5} \cmidrule(lr){6-7} \cmidrule(lr){8-9} \cmidrule(lr){10-11} \cmidrule(lr){12-13} \cmidrule(lr){14-15} \cmidrule(lr){16-17} \cmidrule(lr){18-19}
& F1 & Acc & F1 & Acc & F1 & Acc & F1 & Acc & F1 & Acc & F1 & Acc & F1 & Acc & F1 & Acc & F1 & Acc \\
\midrule
Chinatown & 0.7206 & 0.7259 & 0.9472 & 0.9563 & 0.9714 & 0.9767 & 0.9456 & 0.9592 & 0.9680 & 0.9708 & 0.9712 & 0.9767 & \underline{0.9784} & \underline{0.9825} & 0.9541 & 0.9659 & \textbf{0.9820} & \textbf{0.9839} \\
Melbourne & 0.7186 & 0.7314 & 0.8426 & 0.8421 & \underline{0.8897} & \underline{0.8877} & 0.8030 & 0.8065 & 0.8670 & 0.8655 & 0.8732 & 0.8786 & 0.8876 & 0.8860 & 0.8392 & 0.8364 & \textbf{0.9669} & \textbf{0.9635} \\
PEMS-BAY & 0.6365 & 0.6790 & 0.6712 & 0.6882 & 0.6838 & 0.6929 & 0.6573 & 0.6844 & 0.6736 & 0.6860 & \underline{0.6950} & \underline{0.7055} & 0.6637 & 0.6840 & 0.6770 & 0.6989 & \textbf{0.7091} & \textbf{0.7200} \\
METR-LA & 0.7186 & 0.7314 & 0.7143 & 0.7224 & 0.7295 & 0.7425 & 0.7158 & 0.7269 & 0.7113 & 0.7229 & \underline{0.7483} & \underline{0.7562} & 0.7158 & 0.7266 & 0.7262 & 0.7385 & \textbf{0.7590} & \textbf{0.7684} \\
DodgerLoop & 0.1500 & 0.2153 & 0.3529 & 0.4125 & \underline{0.5435} & \underline{0.5750} & 0.3817 & 0.4250 & 0.5156 & 0.5625 & 0.2442 & 0.3750 & 0.3607 & 0.4500 & 0.1519 & 0.2250 & \textbf{0.5676} & \textbf{0.6000} \\
PEMS-SF & 0.6373 & 0.6503 & 0.7900 & 0.7919 & 0.7468 & 0.7446 & \underline{0.8215} & \textbf{0.8266} & 0.7384 & 0.7514 & 0.7594 & 0.7630 & 0.7857 & 0.7919 & 0.6246 & 0.6705 & \textbf{0.8310} & \underline{0.8227} \\
\midrule
\end{tabular}
\end{table*}

\noindent Then, new features after distribution similarity fusion can be obtained through the similarity measure results, defined as: $\hat{\bm{x}}=(1-\Delta)(\bm{K}^v\bm{x}^v+\bm{K}^g\bm{x}^g+\bm{K}^t\bm{x}^t)+\Delta \bm{x}^v+\Delta\bm{x}^g+\Delta\bm{x}^t$ ($\bm{K}$ represents the training metric of a instance $\bm{x}$). Finally, we use the Multi-Layer Perceptron (MLP) classifier to predict the label of each data, which is realized by minimizing the fusion loss.
Considering that urban traffic profiling is a multi-classification problem, we introduce multi-class cross-entropy loss to calculate the fusion loss, defined as:
\begin{equation}
\mathcal{L}(CE)=-\mathbb{E}_{y \sim\hat{Y}}\sum_{i=1}^{m}y_i log(y_i^{\prime}),
\end{equation}
\noindent where $y_i$ is the real label and $y_i^{\prime}$ represents the probability that the prediction label belongs to category $i$.

The objective loss consists of two parts: the contrastive loss and the fusion loss.  The full loss can be calculated by:
\begin{equation}
\mathcal{L}=\alpha \mathcal{L}(SUP)+ \beta \mathcal{L}(UNS)+ \gamma \mathcal{L}(CE),
\end{equation}
\noindent where $\alpha$, $\beta$, and $\gamma$ are hyperparameters for balancing the influence of different modules.

\section{Experiments}

To comprehensively evaluate the performance of our proposed MTP framework, we conduct extensive experiments on six public time series classification datasets. This section aims to answer the following core research questions (RQs):
\begin{itemize}
    \item \textbf{RQ1:} How is the performance of MTP compared against state-of-the-art baselines?
    \item \textbf{RQ2:} What are the contributions of the core components within MTP to the final performance?
    \item \textbf{RQ3:} How are the multimodal features learned by our model distributed and separated in the feature space?
\end{itemize}

\subsection{Experimental Setting}

\noindent \textbf{Baselines.} Our framework is compared against 8 state-of-the-art
time series models, including TST \cite{zerveas2021tst}, ShapeNet \cite{cheng2021shapenet}, PatchTST \cite{nie2023patchtst}, SVP-T \cite{zhou2023svpt}, LightTS \cite{zhang2023lightts}, ModernTCN \cite{wang2024moderntcn}, CAFO \cite{li2024cafo}, and InterpGN \cite{wen2025interpgn}. These models cover a range of techniques from Transformer-based architectures to shapelet-based methods and pre-training frameworks. For detailed descriptions of these baselines, please refer to Appendix.

\noindent \textbf{Datasets.} Experiments are conducted on six widely-used public benchmarks for time series classification: Chinatown, MelbournePedestrian (Melbourne), PEMS-BAY, METR-LA, DodgerLoopDay (DodgerLoop), and PEMS-SF. Detailed descriptions of all datasets refer to Appendix.

\noindent \textbf{Metrics.} We adopt a set of classification metrics to measure the performance \cite{xu2025nlgt}, including: Accuracy, Macro-Precision, Macro-Recall, and Macro F1-Score.

\noindent \textbf{Implementation Details.} All experiments are implemented using the PyTorch framework and conducted on a single NVIDIA RTX series GPU. To ensure a fair comparison, we follow the optimal parameter settings for each baseline. The detailed parameter configurations for our framework are available in Appendix. All experiments are run 15 times, and we report the average results.

\subsection{RQ1: Performance Comparison}
We first compare the performance of our proposed framework against eight state-of-the-art baselines. The detailed experimental results are presented in Table~\ref{tab:sota_final}. MTP consistently achieves state-of-the-art results, outperforming all baselines in the majority of cases. Specifically, MTP's strength is evident on datasets with clear, classifiable patterns. On the Melbourne dataset, it secures the best performance in both F1-score (0.9669) and Accuracy (0.9635). On the Chinatown dataset, our MTP framework achieves the highest F1-score of 0.9820 and a comparable Accuracy of 0.9839, while PatchTST achieves the second-best results.

MTP continues to excel in large-scale traffic datasets. It achieves the highest F1-score and Accuracy on PEMS-BAY (0.7091 / 0.7200) and METR-LA (0.7590 / 0.7684). Furthermore, on the highly volatile DodgerLoopDay dataset, MTP again ranks first with an F1-score of 0.5676 and an Accuracy of 0.6000, significantly outperforming most baselines. Even on PEMS-SF, MTP's performance is highly competitive, obtaining the second-best Accuracy. These results strongly validate that our framework can learn more comprehensive feature representations, leading to state-of-the-art performance across diverse tasks.

\subsection{RQ2: Ablation Study}

% --- 前导部分 ---
% 请确保您的LaTeX文档前导部分包含了AAAI的官方宏包 (例如 \usepackage{aaai25})
% 以及 \usepackage{multirow} 和 \usepackage{booktabs}

% --- 前导部分 ---
% 请确保您的LaTeX文档前导部分包含了AAAI的官方宏包 (例如 \usepackage{aaai25})
% 以及 \usepackage{multirow} 和 \usepackage{booktabs}

\begin{table}[t!]
\centering
\scriptsize % Use the smallest font to fit in a single column
\setlength{\tabcolsep}{2.5pt} % Reduce space between columns
\caption{Ablation study on two representative datasets. Best results are in \textbf{bold}. MTP is the full model; w/o V: without Visual; w/o T: without Textual; w/o TS: without Timeseries.}
\label{tab:ablation_single_col}
\begin{tabular}{l cccc cccc}
\toprule
\multirow{2}{*}{Variant} & \multicolumn{4}{c}{Melbourne} & \multicolumn{4}{c}{DodgerLoop} \\
\cmidrule(lr){2-5} \cmidrule(lr){6-9}
& Acc & Pre & Rec & F1 & Acc & Pre & Rec & F1 \\
\midrule
\textbf{MTP} & \textbf{0.9672} & \textbf{0.9671} & \textbf{0.9669} & \textbf{0.9669} & \textbf{0.6000} & \textbf{0.6978} & \textbf{0.5923} & \textbf{0.5848} \\
w/o Visual & 0.7593 & 0.7617 & 0.7595 & 0.7584 & 0.2375 & 0.0674 & 0.2348 & 0.1048 \\
w/o Textual & 0.9659 & 0.9660 & 0.9657 & 0.9657 & 0.5375 & 0.6268 & 0.5247 & 0.5315 \\
w/o TS & 0.6839 & 0.6845 & 0.6833 & 0.6766 & 0.5875 & 0.6015 & 0.5788 & 0.5676 \\
\bottomrule
\end{tabular}
\end{table}

We conduct a rigorous ablation study of each component and a sensitivity analysis of key hyperparameters to show the impact of all design components on MTP. 
% We conduct an in-depth analysis of our MTP model, including a rigorous ablation study of each component and a sensitivity analysis of key hyperparameters.

\noindent \textbf{Ablation Study.} We remove all components, including removing the visual branch (`w/o V'), the textual branch (`w/o T'), and the time-series branch (`w/o TS'),  from the full framework to evaluate the impact. The results on two representative datasets are shown in Table~\ref{tab:ablation_single_col}, with full results in the Appendix. Our complete framework consistently outperforms all of its ablated variants. The most pronounced impact is observed on the DodgerLoop dataset upon ablating the visual branch (`w/o V'), where the F1-score drops from 0.5848 to a mere 0.1048. Results on the Melbourne dataset further underscore the contribution of each modality: removing the textual branch (`w/o T') induces a measurable decline in F1-score (from 0.9669 to 0.9657), whereas exclusive reliance on visual and textual modalities without the original time-series (`w/o TS') leads to a substantial performance degradation (F1-score of 0.6766). Thus, it is approved that our proposed modality augmentation and fusion are the core drivers of the model's superior performance.

\begin{figure}[t!] % <-- 关键修改：将 figure* 改为 figure
    \centering
    \begin{subfigure}[b]{0.48\linewidth}
        \includegraphics[width=\linewidth]{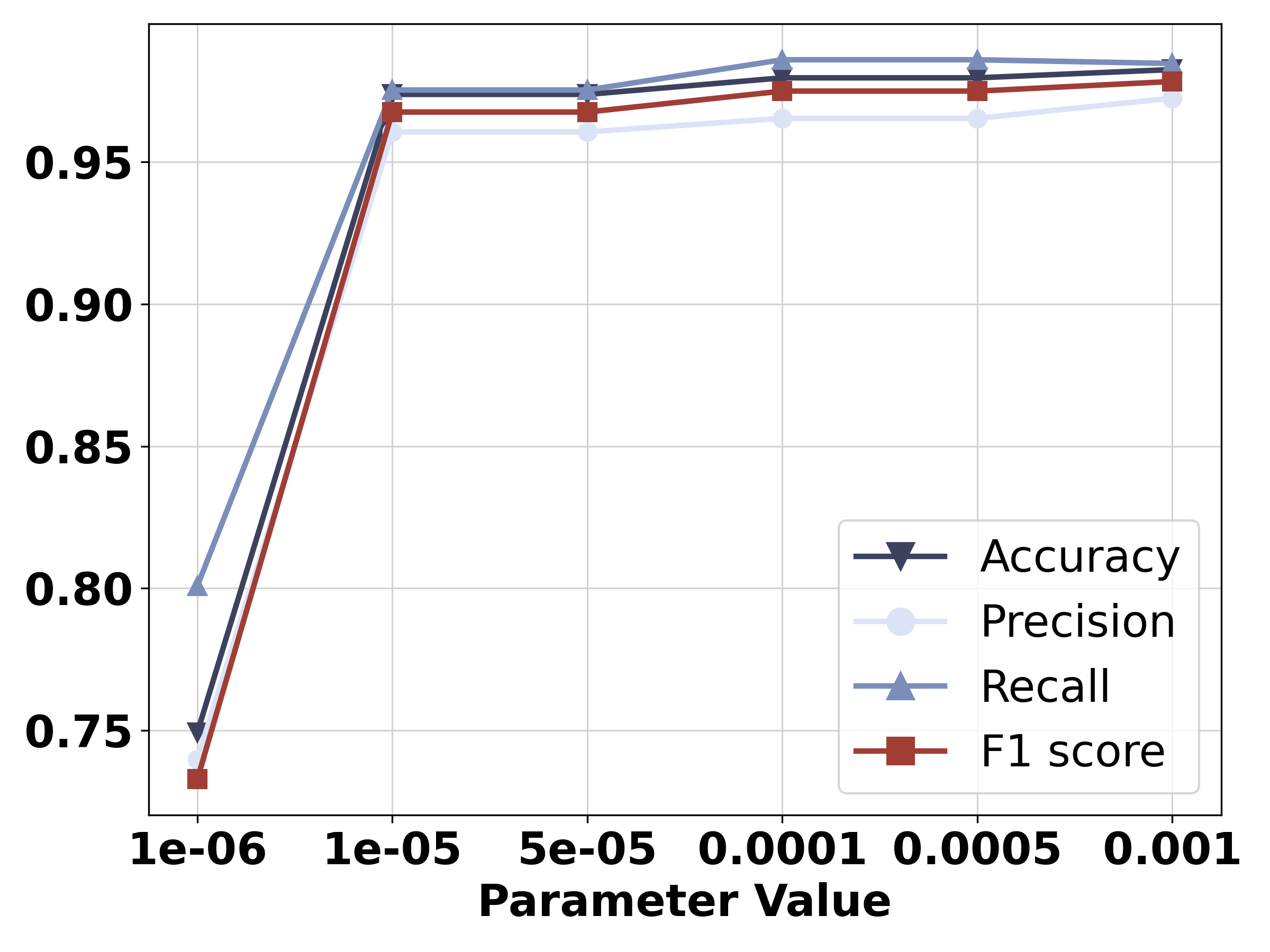}
        \caption{Learning Rate}
        \label{fig:learning_rate}
    \end{subfigure}
    \hfill
    \begin{subfigure}[b]{0.48\linewidth}
        \includegraphics[width=\linewidth]{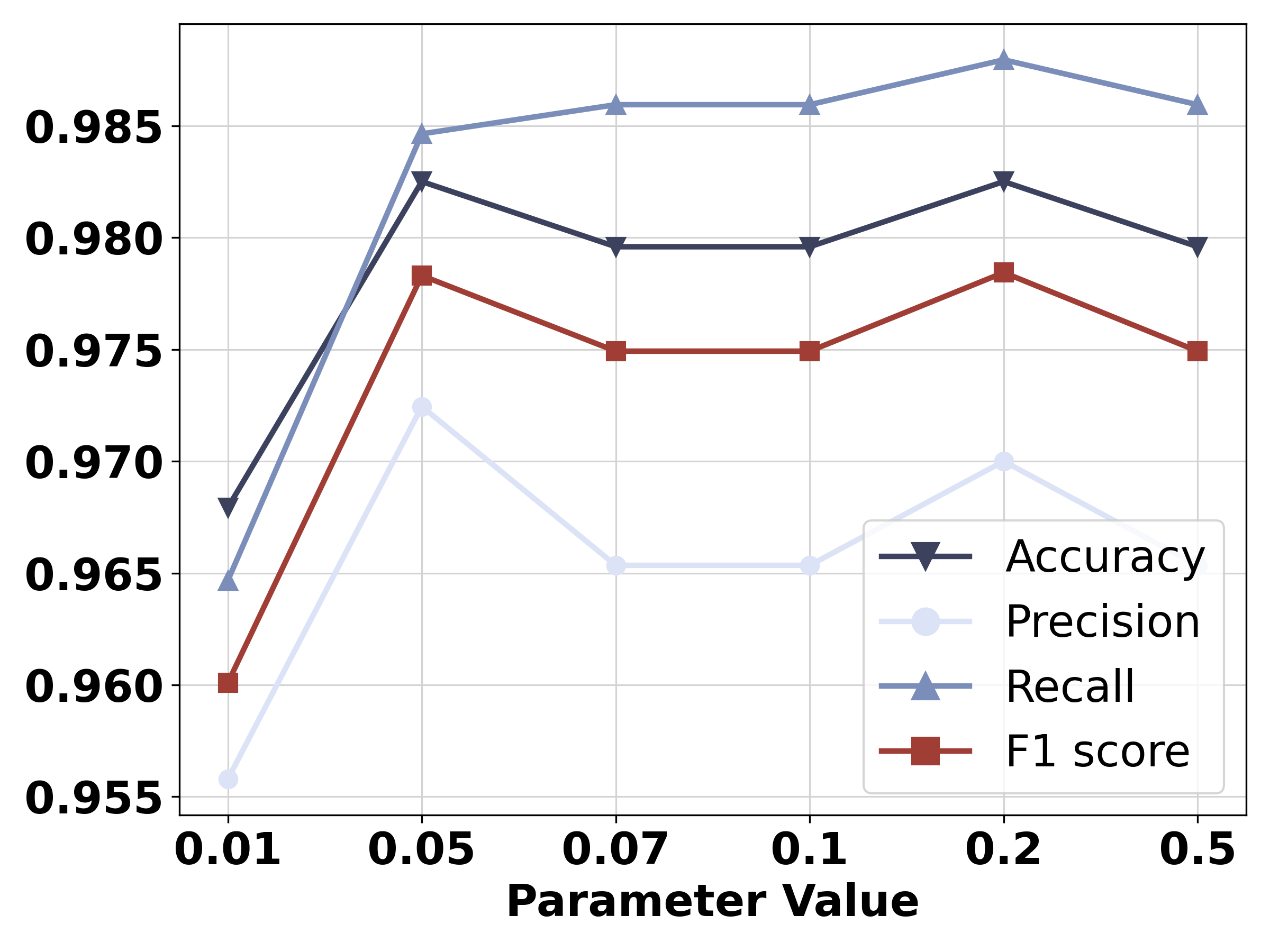}
        \caption{Temperature }
        \label{fig:temperature}
    \end{subfigure}

    % \vspace{2mm} 

    \begin{subfigure}[b]{0.48\linewidth}
        \includegraphics[width=\linewidth]{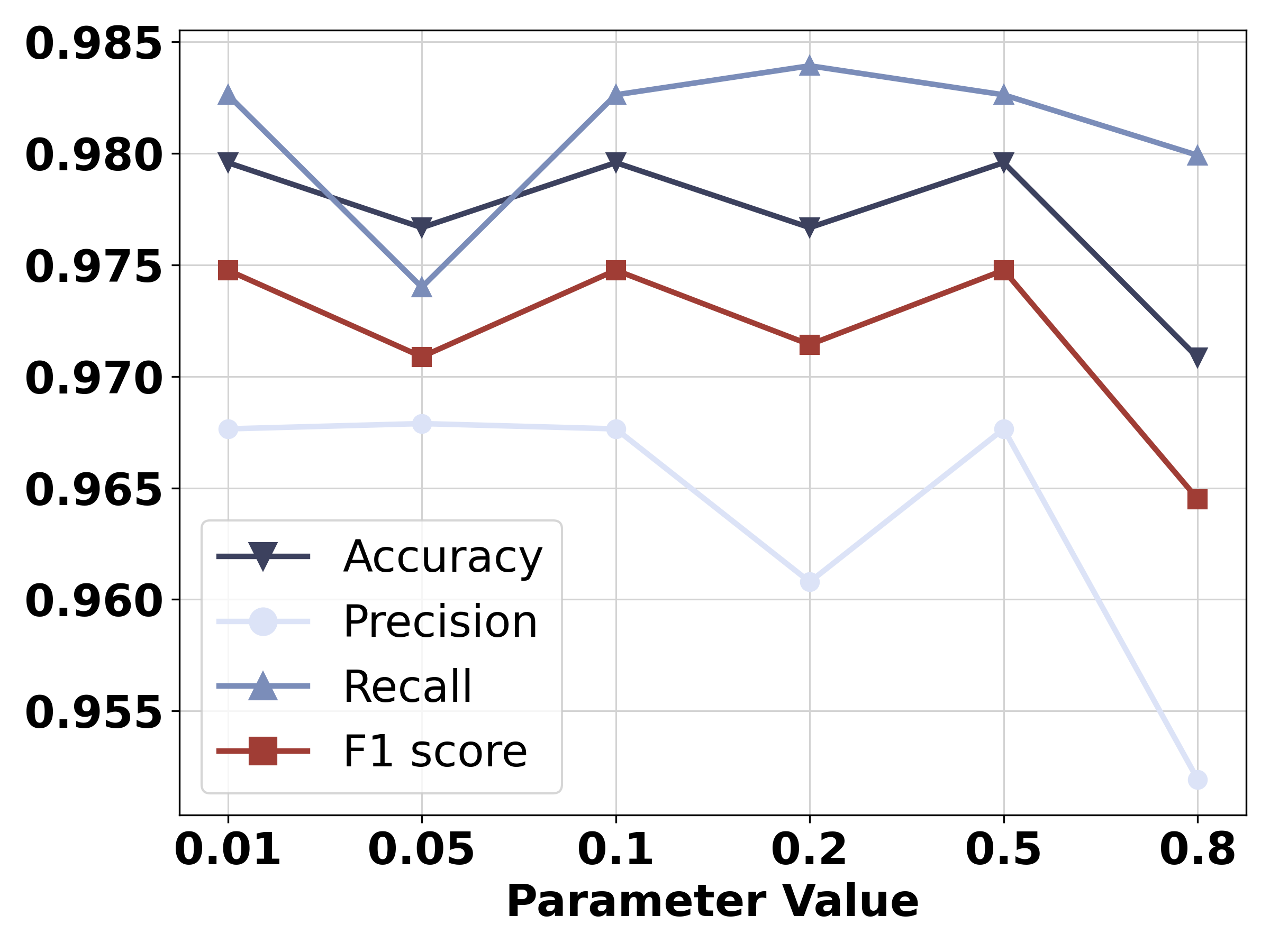}
        \caption{Alpha weight}
        \label{fig:alpha}
    \end{subfigure}
    \hfill
    \begin{subfigure}[b]{0.48\linewidth}
        \includegraphics[width=\linewidth]{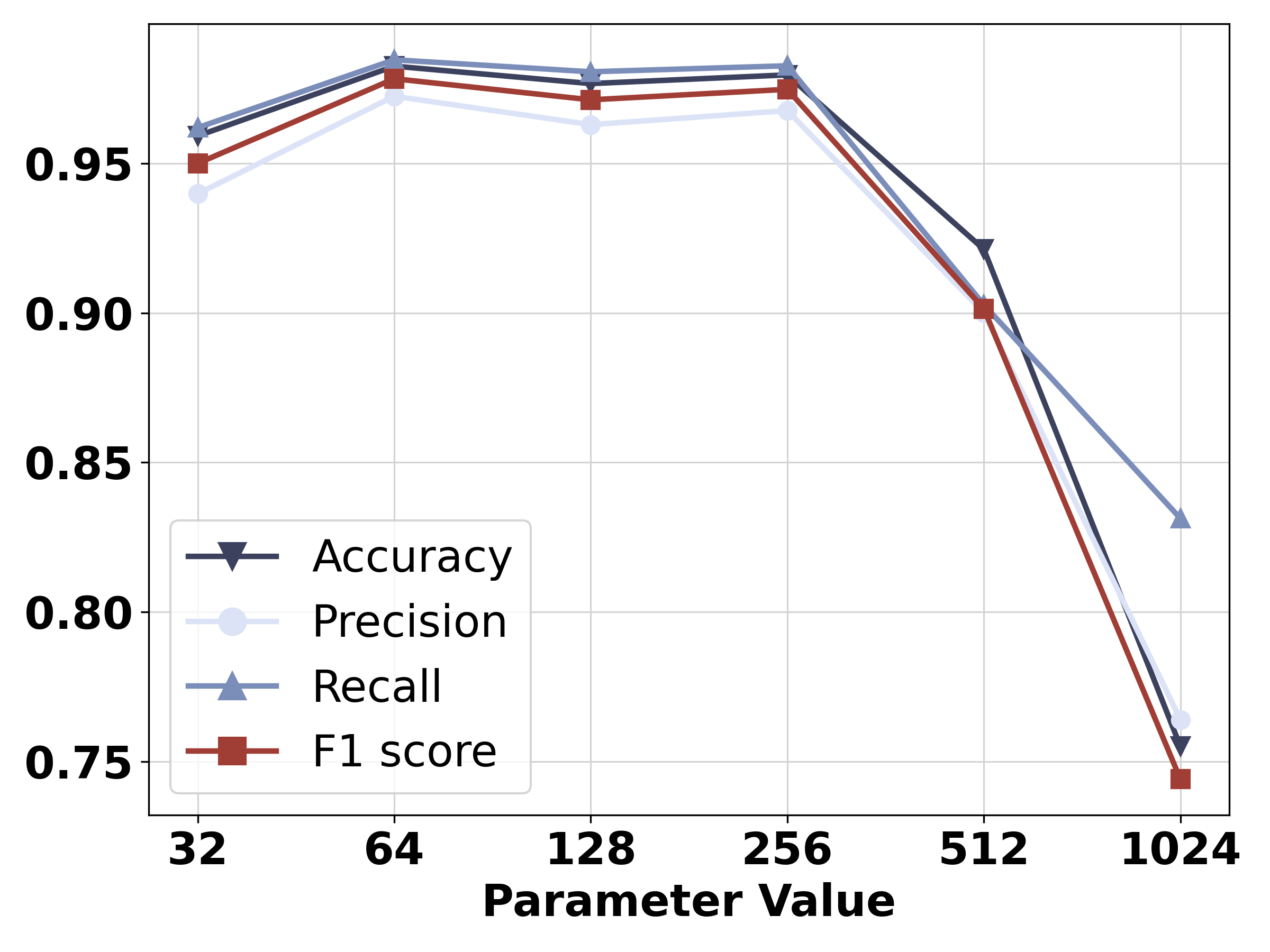}
        \caption{Embedding dimension}
        \label{fig:d_model}
    \end{subfigure}
    
    \caption{Hyperparameter sensitivity analysis on four key parameters: (a) Learning Rate, (b) Temperature, (c) Alpha weight, and (d) Embedding dimension.}
    \label{fig:hyperparameter_sensitivity}
\end{figure} % <-- 关键修改：将 figure* 改为 figure

% --- Figure (METR-LA) - Single Column, 2x2 Grid Layout ---
\begin{figure}[t!] % <-- 关键修改：将 figure* 改为 figure
    \centering
    \begin{subfigure}[b]{0.43\linewidth}
        \includegraphics[width=\linewidth]{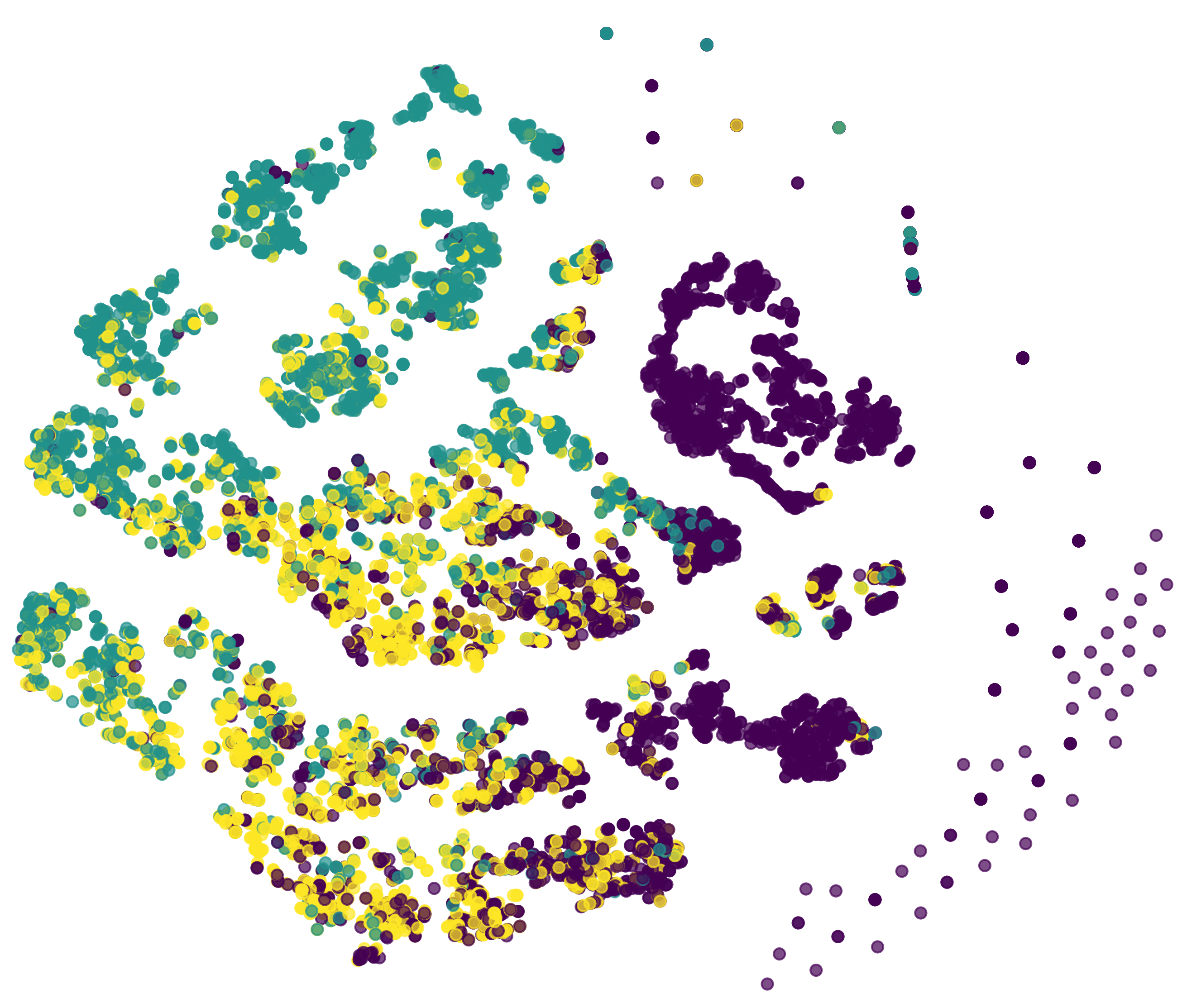}
        \caption{Final fused features}
        \label{fig:metr-la-fused}
    \end{subfigure}
    \hfill
    \begin{subfigure}[b]{0.43\linewidth}
        \includegraphics[width=\linewidth]{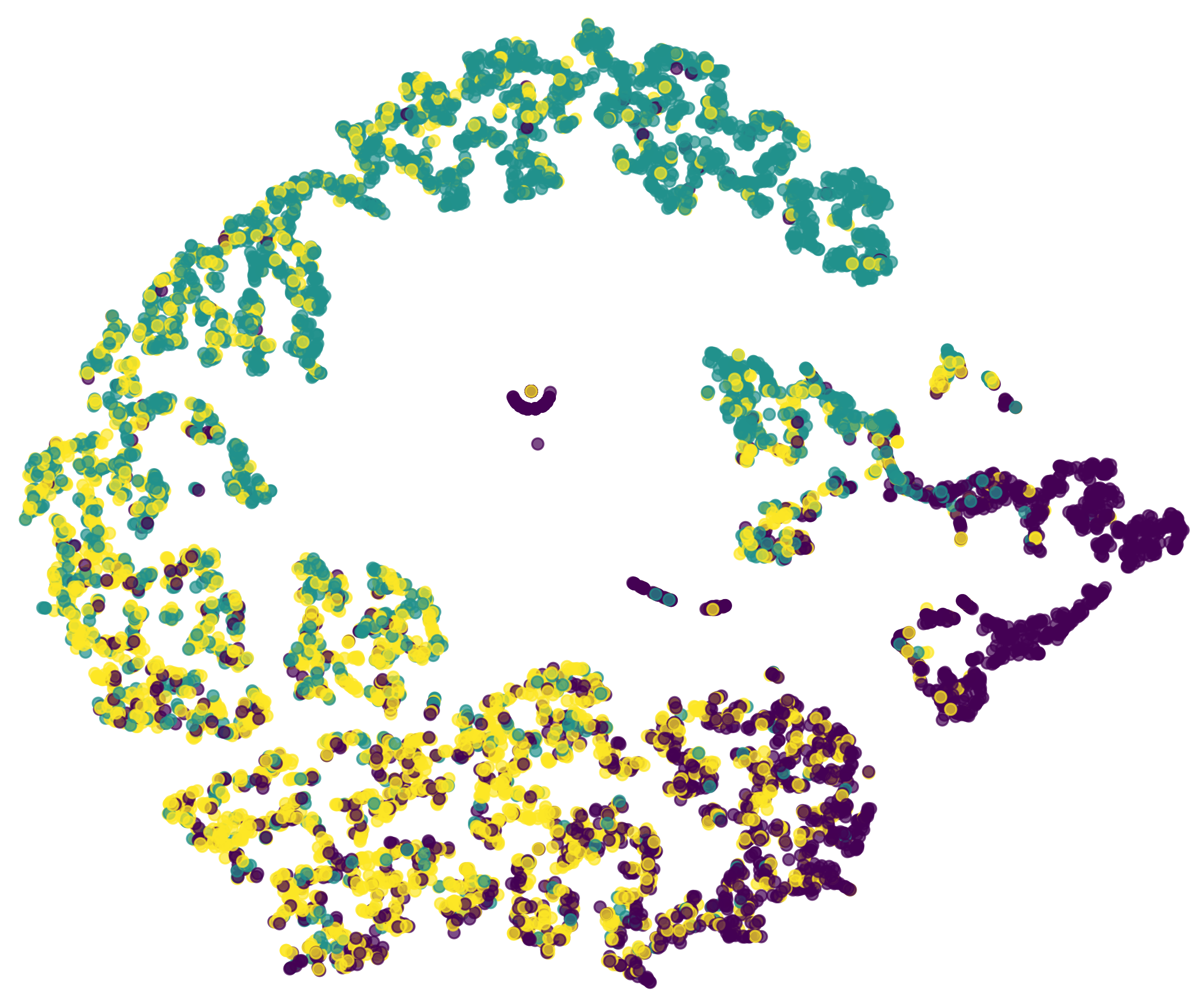}
        \caption{Image features}
        \label{fig:metr-la-image}
    \end{subfigure}

    \begin{subfigure}[b]{0.43\linewidth}
        \includegraphics[width=\linewidth]{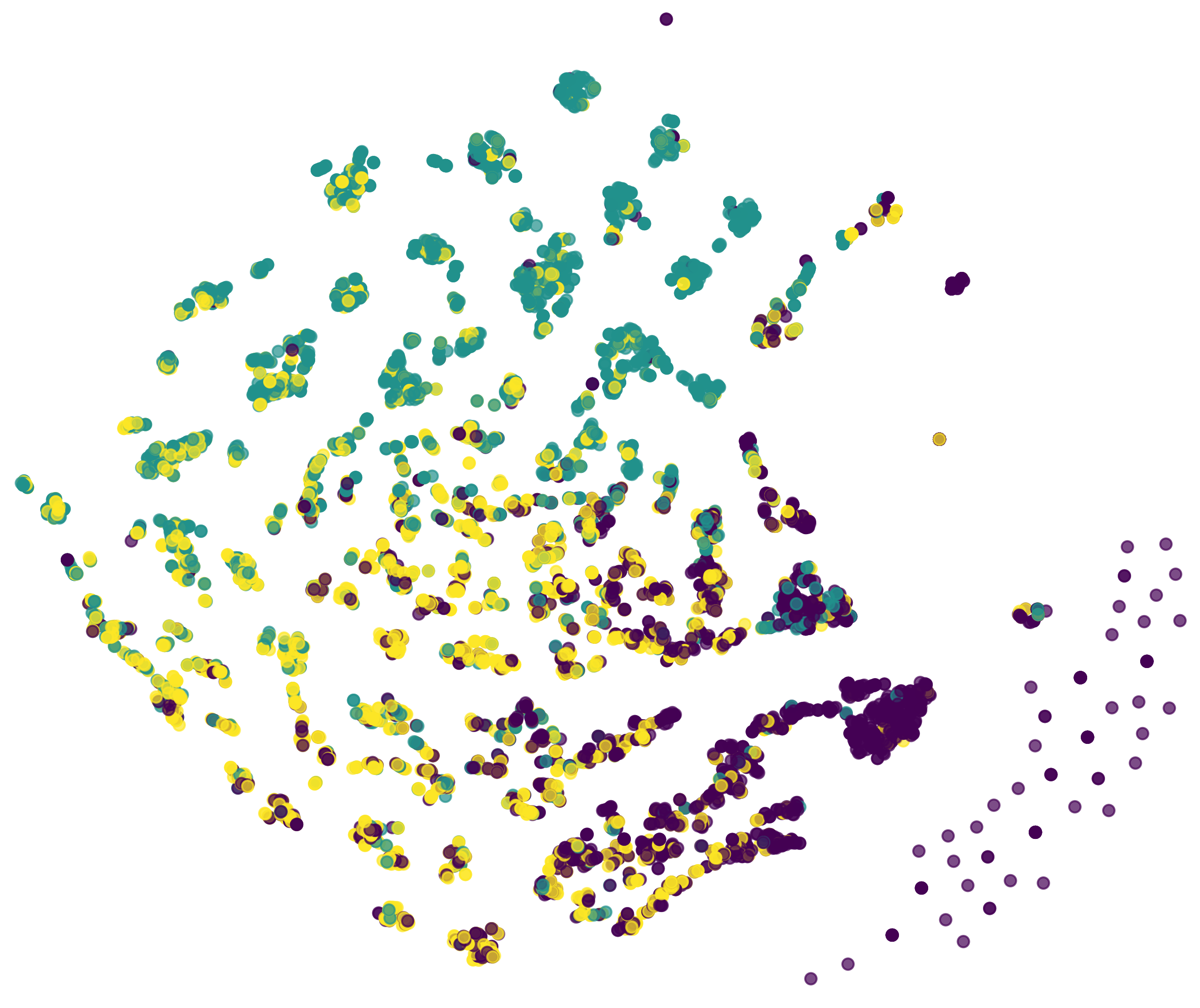}
        \caption{Text features}
        \label{fig:metr-la-text}
    \end{subfigure}
    \hfill
    \begin{subfigure}[b]{0.43\linewidth}
        \includegraphics[width=\linewidth]{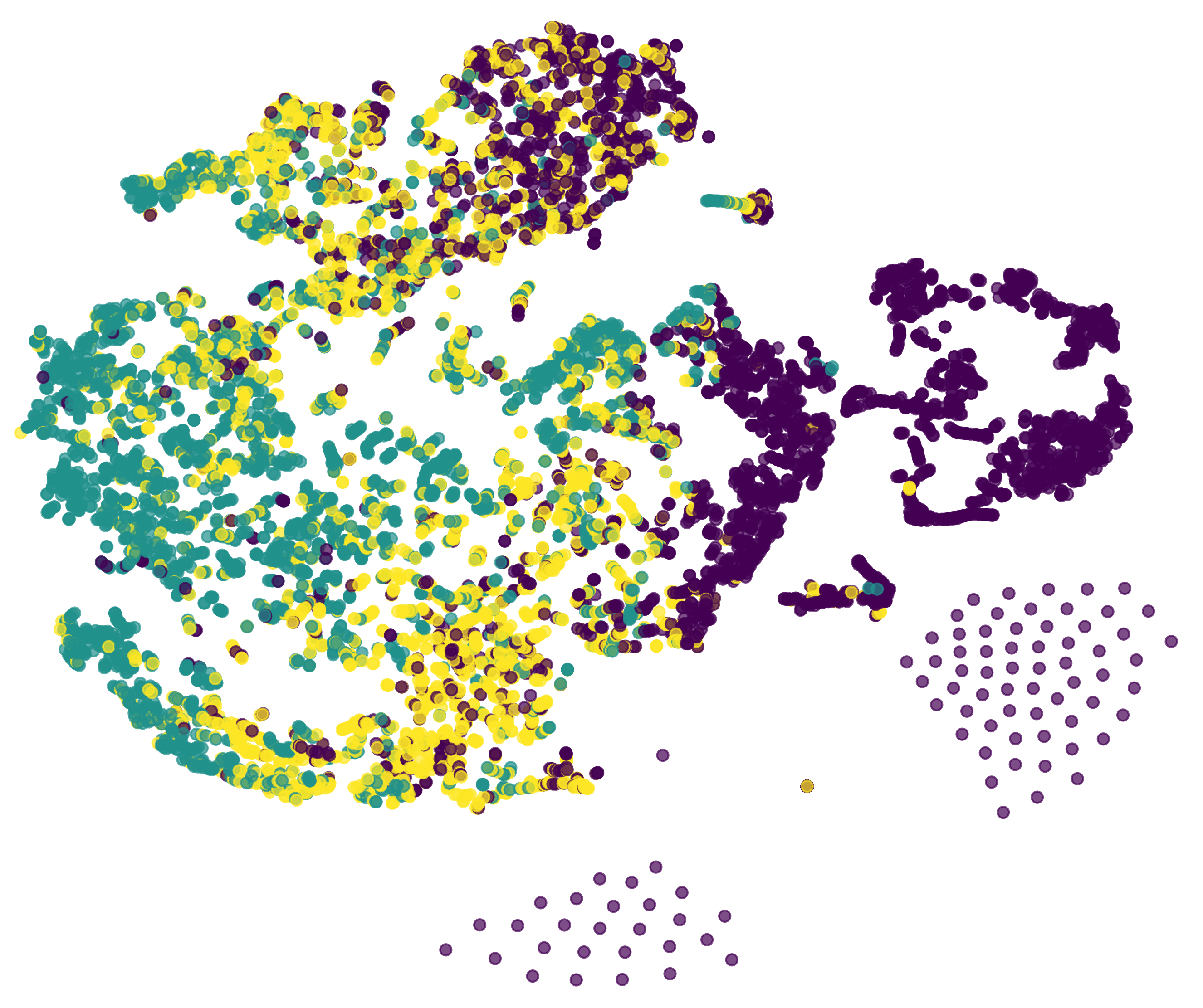}
        \caption{Time series features}
        \label{fig:metr-la-timeseries}
    \end{subfigure}
    
    \caption{Comparative t-SNE visualizations on the METR-LA dataset, which contains three types of labels.}
    \label{fig:tsne-metr-la}
\end{figure} % <-- 关键修改：将 figure* 改为 figure

\noindent \textbf{Hyperparameter Sensitivity Analysis.} We investigate the sensitivity of MTP to four key hyperparameters, shown in Figure~\ref{fig:hyperparameter_sensitivity}. The results indicate that our framework exhibits robustness across different hyperparameters. For the learning rate, performance reaches its peak around $1e-4$. Regarding the temperature parameter, the optimal range is between 0.05 and 0.1. MTP demonstrates a distinct preference for a smaller alpha value in contrastive loss, with the best performance achieved at 0.1. Finally, for the embedding dimension, performance plateaus once the dimension reaches 128. These findings collectively validate that MTP is stable and does not necessitate extensive hyperparameter tuning.

% The analysis reveals that our model is robust to a reasonable range of settings. For the Learning Rate, performance peaks around $1e-4$. For the contrastive loss Temperature, the optimal range is between 0.05 and 0.1. The model shows a clear preference for a smaller Alpha (contrastive loss weight), with performance being best at 0.1. Finally, for the Embedding Dimension, performance plateaus after 128. This analysis demonstrates that MTP is stable and does not require extensive tuning.

\subsection{RQ3: Qualitative Analysis}
To intuitively understand the effectiveness of our framework, we use t-SNE to visualize the feature distributions on the METR-LA test set, as shown in Figure~\ref{fig:tsne-metr-la}. the final fused features learned by our complete framework form highly cohesive and clearly separated clusters in the 2D space. Samples from different classes represented by different colors are distinctly separated with minimal overlap. This provides strong visual evidence for the high classification performance reported in Table~\ref{tab:sota_final}. In contrast, the feature distributions from single modalities are more diffuse and intermingled. This qualitative result is consistent with our conclusions from the ablation study, demonstrating that our fusion module successfully integrates complementary information to produce a more powerful and discriminative final representation. For more visualizations on other datasets, please refer to the Appendix.

\section{Conclusion}
In this paper, we propose a novel multi-modal urban traffic profiling framework, called MTP, which addresses the issue that existing traffic profiling methods rely on a single numerical modality and overlook the semantic information in multi-modal heterogeneous data. MTP learns multimodal features in the frequency domain from three perspectives: numerical, visual, and textual. Specifically, MTP processes numerical information using frequency multi-layer perceptions, performs visual augmentation by converting raw data into periodic images and frequency images, and generates descriptive texts based on information such as topics and backgrounds for textual augmentation. Besides, MTP designs hierarchical contrastive learning to fuse the three modalities. Experiments on six real-world datasets demonstrate that our framework significantly outperforms state-of-the-art methods. Future work will involve integrating more types of urban modal data and exploring fine-grained modeling mechanisms for cross-modal correlations.

\section{Acknowledgments}
This work was supported in part by the National Natural Science Foundation of China under Grant 62372242, Jiangsu Provincial Major Project on Basic Research of Cutting-edge and Leading Technologies, under grant no. BK20232032, and Open Research Projects of the State Key Laboratory for New Technologies of Computer Software, Nanjing University, under grant no. KFKT2025B64.

\bibliography{aaai2026}

% ===================================================================
% Appendix
% ===================================================================
\newpage

\appendix

\section{Appendix A}
\label{sec:appendix_baselines}
\paragraph{Baselines: } 
For a comprehensive performance comparison, we compare our framework against the following 8 state-of-the-art time series models.
\begin{itemize}
    \item \textbf{TST (Time Series Transformer) \cite{zerveas2021tst}:} Directly applies the standard Transformer encoder architecture to the spatial domain. It utilizes a self-attention mechanism to capture pairwise relationships across all time steps.
    \item \textbf{ShapeNet \cite{cheng2021shapenet}:} A shapelet-based neural network for multivariate time series classification. It learns discriminative shapelets and feeds the extracted shapelet features into a fully connected network to capture both local patterns and global dependencies.
    \item \textbf{PatchTST \cite{nie2023patchtst}:} A recent Transformer-based model that treats a time series as a sequence of patches (subseries). It processes each channel independently, enabling effective representation learning for long-term forecasting and classification tasks.
    \item \textbf{LightTS \cite{zhang2023lightts}:} A lightweight framework designed for time series classification. It employs an adaptive integrated distillation technique to transfer knowledge from multiple heterogeneous teacher models into a single lightweight student model.
    \item \textbf{SVP-T \cite{zhou2023svpt}:} A pre-training framework for time series data that operates on two levels. It learns representations from both the shape-level (local patterns) and velocity-level (trend information) of the time series, aiming to create more robust features for downstream tasks.
    \item \textbf{ModernTCN \cite{wang2024moderntcn}:} A modernized version of the classic Temporal Convolutional Network (TCN). It incorporates modern CNN design principles, such as depthwise separable convolutions, to enhance model performance and scalability.
    \item \textbf{CAFO \cite{li2024cafo}:} This model is a convolutional attention-based backbone network designed for time series classification tasks. It effectively combines the local feature extraction capabilities of convolutional layers with the ability of attention mechanisms to capture long-range dependencies.
    \item \textbf{InterpGN \cite{wen2025interpgn}:} A framework aiming to combine model performance with interpretability for time series classification. It uses learnable shapelets as an interpretable module and fuses its output with a powerful "black-box" network via a gated mechanism to provide both accurate and explainable predictions.
\end{itemize}

\section{Appendix B}
\label{sec:appendix_datasets}
\paragraph{Datasets: } 
We conducted all our experiments on the following six public benchmark datasets for time series classification:
\begin{itemize}
    \item \textbf{Chinatown\footnote{\url{https://www.timeseriesclassification.com/description.php?Dataset=Chinatown}}:} Originating from the UCR/UEA Time Series Classification Archive. This dataset records the number of pedestrians at two different locations in a Chinatown district, making it a multivariate time series classification task.
    \item \textbf{METR-LA\footnote{\url{https://zenodo.org/records/5146275/files/METR-LA.csv}}:} A large-scale traffic dataset containing traffic speed data from 207 sensors on Los Angeles County highways over 4 months. It is a classic multivariate time series dataset widely used for traffic flow and speed forecasting research.
    \item \textbf{MelbournePedestrian\footnote{\url{https://zenodo.org/records/4656626/files/pedestrian_counts_dataset.zip}}:} Contains hourly pedestrian counts from 2015 to 2017 at various locations in the city center of Melbourne, Australia. This dataset is used for urban mobility pattern analysis and pedestrian flow prediction.
    \item \textbf{PEMS-BAY\footnote{\url{https://zenodo.org/records/5146275/files/PEMS-BAY.csv}}:} Another large-scale traffic dataset from the California Department of Transportation's Performance Measurement System (PeMS). It includes traffic data from 325 sensors in the San Francisco Bay Area over 6 months.
    \item \textbf{DodgerLoopDay\footnote{\url{https://www.timeseriesclassification.com/description.php?Dataset=DodgerLoopDay}}:} Also from the UCR/UEA Time Series Classification Archive, this dataset contains counts of vehicles on a road leading to the Dodgers Stadium in Los Angeles. The classification task is to distinguish between game days and non-game days based on traffic patterns.
    \item \textbf{PEMS-SF\footnote{\url{https://www.timeseriesclassification.com/description.php?Dataset=PEMS-SF}}:} Collected by the Caltrans Performance Measurement System (PeMS), this dataset contains traffic occupancy rates from 963 sensors on the highways of the San Francisco Bay Area and is a widely used benchmark for traffic classification.
\end{itemize}

\section{Appendix C}
\label{sec:appendix_implementation}
\paragraph{Implementation Details:} 
Our framework and all comparison experiments are implemented using the PyTorch framework and executed on a single NVIDIA RTX 3090 GPU. For the modality generation, images are created with a size of 64x64 pixels, and the maximum text length is set to 128 tokens. The model is trained for 50 epochs with a batch size of 64. We utilize the AdamW optimizer with an initial learning rate of 1e-4, a weight decay of 0.01, and a learning rate scheduler with a linear warmup. For the loss function, the contrastive loss weight ($\alpha$) is set to 0.1, and the temperature coefficient ($\tau$) is 0.07. A five-fold cross-validation approach is employed for training to ensure the robustness of the results. The performance is evaluated using Accuracy, Precision, Recall, and F1 score metrics. All experiments are run 15 times, and we report the arithmetic average as the final result.
% ===================================================================
% 附录 D: 消融实验详细结果
% ===================================================================

\begin{table}[t!]
\centering
\scriptsize % 使用最小字体以适应单栏宽度
\caption{Detailed ablation study results for the remaining four datasets. The best result in each row is in \textbf{bold}.}
\label{tab:ablation_appendix}
% 使用 tabular* 环境，宽度设为 \columnwidth (当前栏的宽度)
% @{\extraclolsep{\fill}} 会自动在列间添加空白，使表格撑满指定宽度
\begin{tabular*}{\columnwidth}{@{\extracolsep{\fill}} ll cccc}
\toprule
Dataset & Metric & \textbf{MTP} & w/o V & w/o T & w/o TS \\
\midrule
\multirow{4}{*}{Chinatown} & Accuracy & \textbf{0.9854} & 0.9271 & 0.9796 & 0.9563 \\
& Precision & \textbf{0.9747} & 0.9008 & 0.9653 & 0.9312 \\
& Recall & \textbf{0.9900} & 0.9233 & 0.9859 & 0.9699 \\
& F1-score & \textbf{0.9820} & 0.9110 & 0.9749 & 0.9475 \\
\midrule
\multirow{4}{*}{PEMS-BAY} & Accuracy & \textbf{0.7128} & 0.7051 & 0.7071 & 0.6639 \\
& Precision & \textbf{0.7133} & 0.6929 & 0.7037 & 0.6606 \\
& Recall & \textbf{0.7093} & 0.6916 & 0.7066 & 0.6474 \\
& F1-score & \textbf{0.7091} & 0.6905 & 0.7050 & 0.6478 \\
\midrule
\multirow{4}{*}{METR-LA} & Accuracy & \textbf{0.7680} & 0.7623 & 0.7671 & 0.7342 \\
& Precision & \textbf{0.7592} & 0.7584 & 0.7590 & 0.7304 \\
& Recall & \textbf{0.7592} & 0.7543 & 0.7520 & 0.7235 \\
& F1-score & \textbf{0.7590} & 0.7552 & 0.7526 & 0.7248 \\
\midrule
\multirow{4}{*}{PEMS-SF} & Accuracy & \textbf{0.7977} & 0.6127 & 0.5549 & 0.6185 \\
& Precision & \textbf{0.8100} & 0.5967 & 0.5641 & 0.6035 \\
& Recall & \textbf{0.7912} & 0.5999 & 0.5448 & 0.6114 \\
& F1-score & \textbf{0.7888} & 0.5810 & 0.5428 & 0.5997 \\
\bottomrule
\end{tabular*}
\end{table}

\begin{figure}[t!] 
    \centering
    \begin{subfigure}[b]{0.48\linewidth}
        \includegraphics[width=\linewidth]{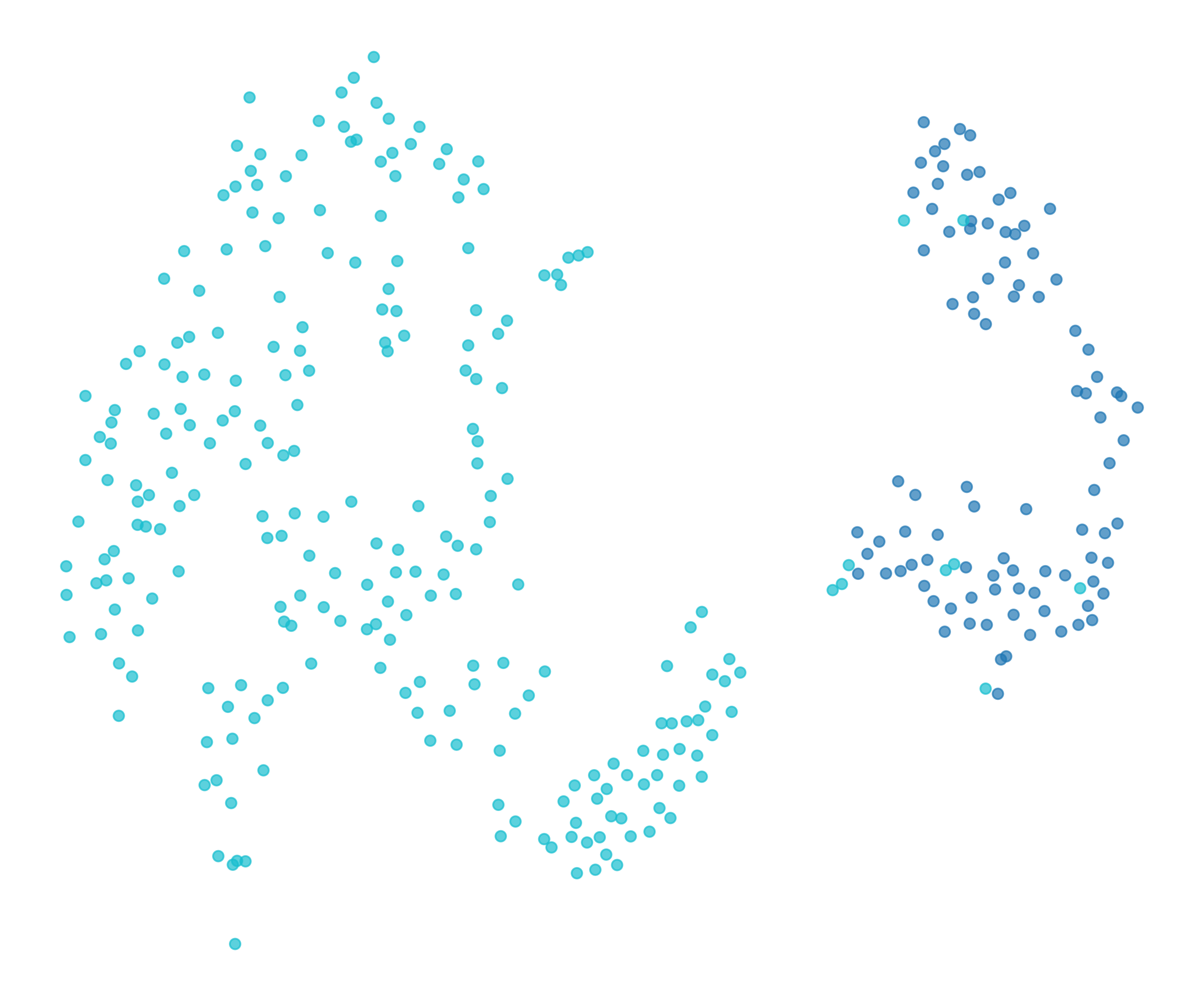}
        \caption{Final fused features}
        \label{fig:chinatown-fused-appendix}
    \end{subfigure}
    \hfill
    \begin{subfigure}[b]{0.48\linewidth}
        \includegraphics[width=\linewidth]{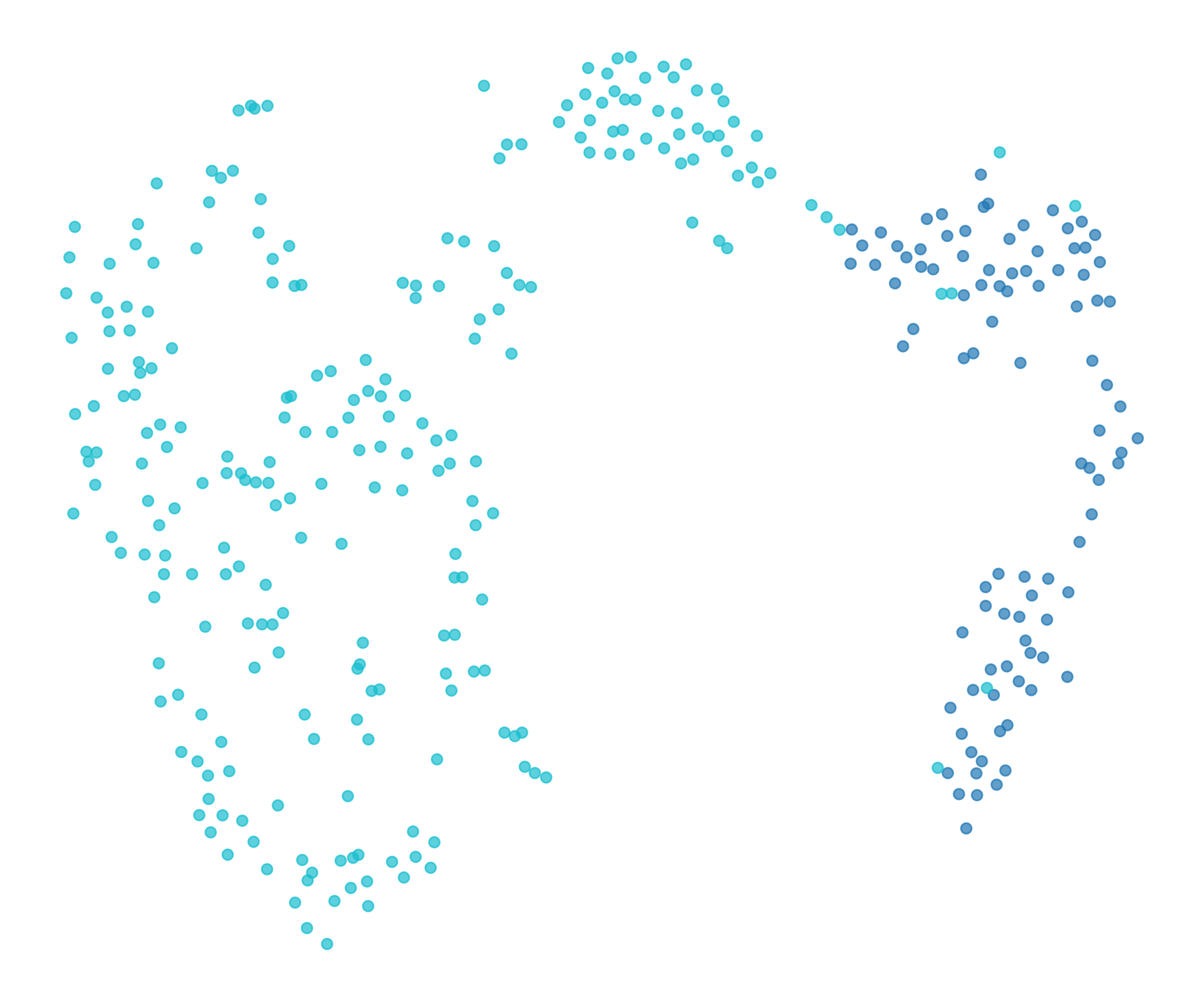}
        \caption{Image features}
        \label{fig:chinatown-image-appendix}
    \end{subfigure}
    
    \vspace{2mm} 

    \begin{subfigure}[b]{0.48\linewidth}
        \includegraphics[width=\linewidth]{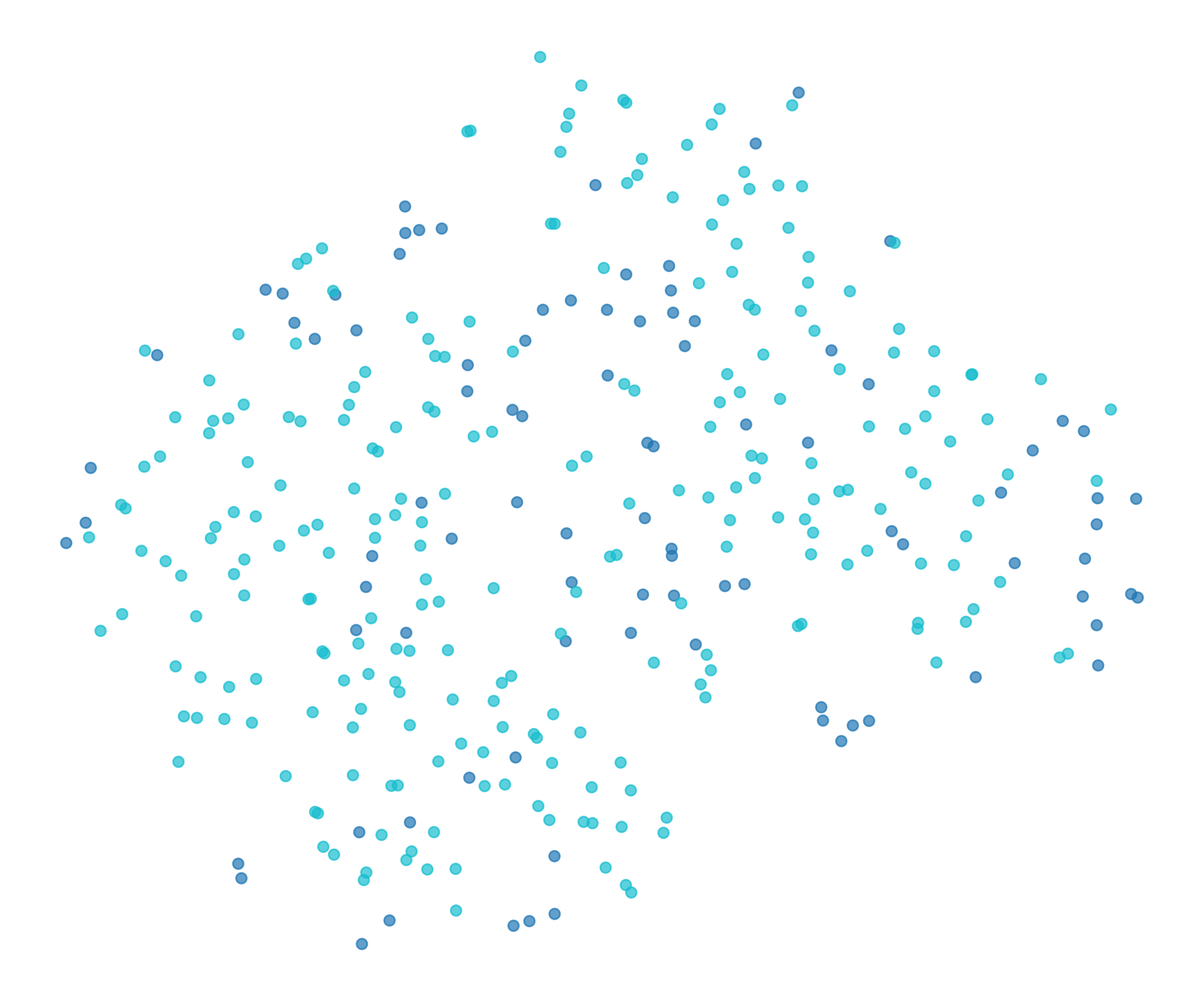}
        \caption{Text features}
        \label{fig:chinatown-text-appendix}
    \end{subfigure}
    \hfill
    \begin{subfigure}[b]{0.48\linewidth}
        \includegraphics[width=\linewidth]{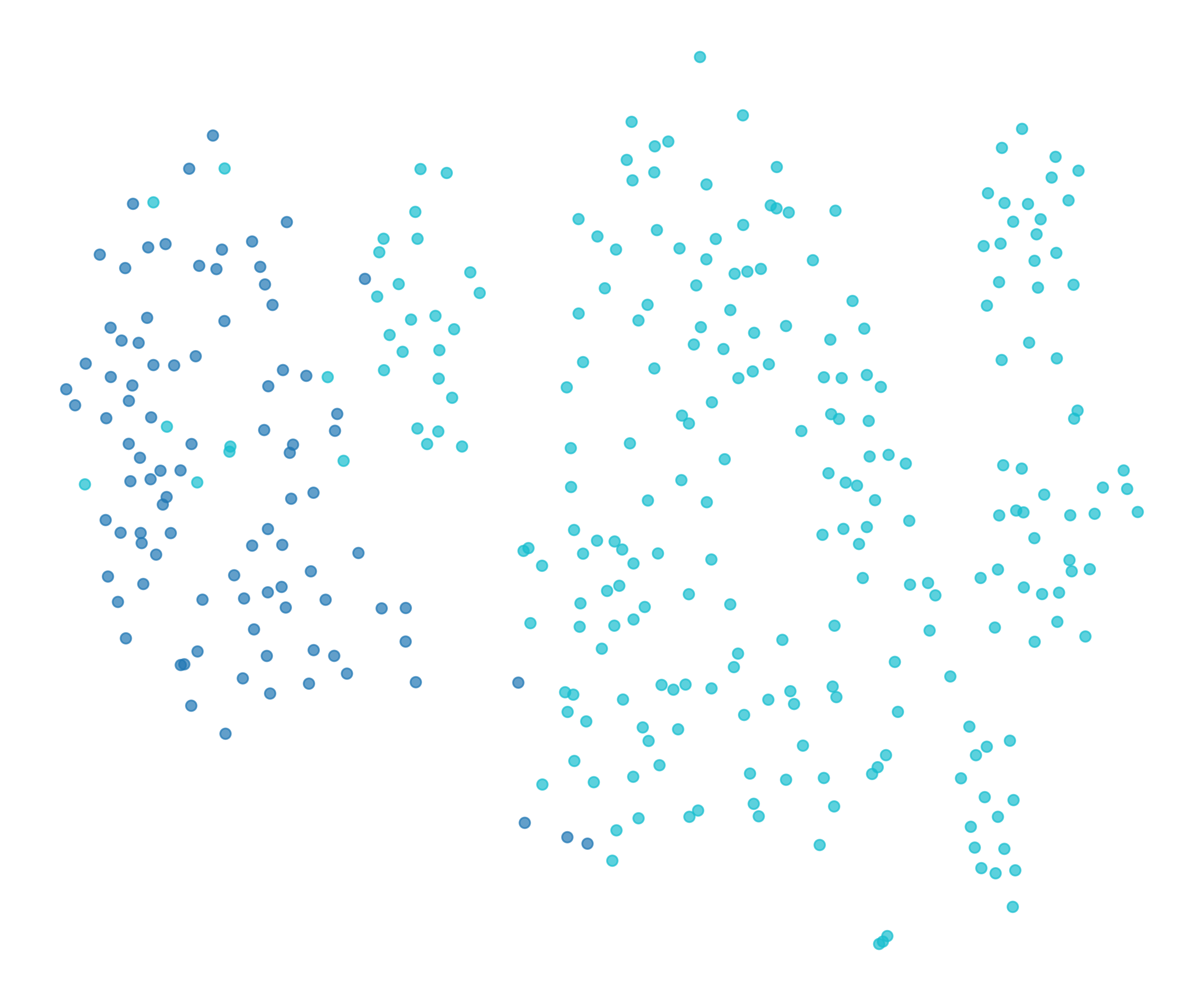}
        \caption{Time series features}
        \label{fig:chinatown-timeseries-appendix}
    \end{subfigure}
    
    \caption{Comparative t-SNE visualizations on the Chinatown dataset. Color coding: blue (Class 0), green (Class 1).}
    \label{fig:tsne-chinatown-appendix}
\end{figure}

\section{Appendix D}
\label{sec:appendix_ablation}
\paragraph{Detailed Ablation Study Results:} 
The detailed results in Table~\ref{tab:ablation_appendix} are consistent with the findings presented in the main paper. Our complete MTP framework demonstrates the best or highly competitive performance across nearly all metrics on these remaining datasets. For instance, on the PEMS-SF dataset, removing the visual modality (`w/o V') results in a significant drop in F1-score from 0.7888 to 0.5810. On the METR-LA dataset, while the performance of all variants is notably close, the full framework still maintains a slight edge, showcasing its robustness. These unabridged results provide further evidence supporting the conclusions drawn from our ablation study.

% ===================================================================
% 附录 E: 补充可视化结果
% ===================================================================
\section{Appendix E}
\label{sec:appendix_qualitative}
\paragraph{Additional Qualitative Results:}
The results are shown in Figure~\ref{fig:tsne-chinatown-appendix}. Consistent with the visualizations on the METR-LA dataset presented in the main paper, the fused features (a) learned by MTP on the Chinatown dataset form clear and well-separated clusters. In contrast, the features from the individual modalities show substantially more overlap between classes. These results provide additional qualitative evidence that our multimodal fusion mechanism enhances the discriminative power of the learned representations, thereby improving classification performance.

\section{Appendix F}
\label{appendix:f}
\subsection{Data Preprocessing Details}

\noindent\textbf{Label Generation Strategy for METR-LA Dataset:} \\
To enable the quantitative classification of complex traffic dynamics, we preprocess the continuous speed recordings from the METR-LA dataset into discrete traffic states. Our classification framework is conceptually grounded in the Level of Service (LOS) standards defined by the \textit{Highway Capacity Manual} (HCM). We consolidate the six standard LOS grades (A through F) into three distinct categories, including``Low Congestion,'' ``Moderate Congestion,'' and ``High Congestion'', to adapt to the machine learning classification task.

The specific numerical thresholds are determined using the ``Percentage of Free-Flow Speed'' (FFS) method. Given that the average Free-Flow Speed on the highways monitored in the METR-LA dataset is approximately 65 mph, we define the traffic state labels based on the following speed thresholds:

\begin{itemize}
    \item \textbf{High Congestion (Traffic Breakdown):} Speed $< 40$ mph (approximately $<60\%$ of FFS). This state corresponds to LOS E/F, indicating severe congestion and stop-and-go traffic conditions.
    \item \textbf{Moderate Congestion (Transitional Flow):} Speed $\in [40, 60]$ mph. This state corresponds to LOS C/D, representing a transitional phase where traffic flow becomes unstable but has not yet fully broken down.
    \item \textbf{Low Congestion (Free Flow):} Speed $> 60$ mph. This state corresponds to LOS A/B, indicating smooth and uninhibited traffic flow.
\end{itemize}

This physics-informed labeling strategy ensures that the classification targets are interpretable and consistent with established transportation engineering principles.

\end{document}